\documentclass[10pt,twocolumn,letterpaper]{article}

\usepackage{iccv}
\usepackage{times}
\usepackage{epsfig}
\usepackage{graphicx}
\usepackage{amsmath}
\usepackage{amssymb}
\usepackage{caption}
\usepackage{multirow}
\usepackage{subcaption}
\usepackage[ruled,vlined]{algorithm2e}
\newcommand*\samethanks[1][\value{footnote}]{\footnotemark[#1]}

\usepackage[pagebackref=true,breaklinks=true,letterpaper=true,colorlinks,bookmarks=false]{hyperref}

\iccvfinalcopy 

\def\iccvPaperID{xxxx} 

\ificcvfinal\fi

\begin{document}

\title{Discovering A Variety of Objects in Spatio-Temporal Human-Object Interactions\thanks{Technical report. A part of the \href{http://hake-mvig.cn/}{HAKE} project. HAKE also provides human body Part States (PaSta) labels~\cite{pastanet} paired with DIO on AVA.}}

\author{
Yong-Lu Li$^{1}$\thanks{The first two authors contribute equally.},~
Hongwei Fan$^{2}$\samethanks,~
Zuoyu Qiu$^{1}$,~
Yiming Dou$^{1}$,~
Liang Xu$^{1}$,~
Hao-Shu Fang$^{1}$,\\
Peiyang Guo$^{1}$,~
Haisheng Su$^{2}$,~
Dongliang Wang$^{2}$,~
Wei Wu$^{2}$,~
Cewu Lu$^{1}$\thanks{Corresponding author.}\\
$^{1}$Shanghai Jiao Tong University,
$^{2}$Sensetime\\
\tt\small{\{yonglu\_li, 17803091056, douyiming, lucewu\}@sjtu.edu.cn}\\
\tt\small{\{fanhongwei1, suhaisheng, wangdongliang, wuwei\}@senseauto.com, \{liangxuy96, fhaoshu\}@gmail.com}\\
\tt\small{\{hellobrian18\}@163.com}
}

\maketitle
\ificcvfinal\thispagestyle{empty}\fi

\begin{abstract}
Spatio-temporal Human-Object Interaction (ST-HOI) detection aims at detecting HOIs from videos, which is crucial for activity understanding.
In daily HOIs, humans often interact with a variety of objects, \eg, \textit{holding} and \textit{touching} dozens of household items in cleaning. However, existing whole body-object interaction video benchmarks usually provide limited object classes. Here, we introduce a new benchmark based on AVA: \textbf{D}iscovering \textbf{I}nteracted \textbf{O}bjects (\textbf{DIO}) including 51 interactions and \textbf{1,000}+ objects. 
Accordingly, an ST-HOI learning task is proposed expecting vision systems to track human actors, detect interactions, and simultaneously discover interacted objects. Even though today's detectors/trackers excel in object detection/tracking tasks, they cannot localize diverse/unseen objects in DIO well. This profoundly reveals the limitation of current vision systems and poses a great challenge.
Thus, how to leverage spatio-temporal cues to address object discovery is explored, and a baseline \textbf{H}ierarchical \textbf{P}robe \textbf{N}etwork (\textbf{HPN}) is devised to discover interacted objects utilizing hierarchical spatio-temporal human/context cues.
In extensive experiments, HPN demonstrates decent performance. 
\textbf{Data and code are available at \url{https://github.com/DirtyHarryLYL/HAKE-AVA}
}.
\end{abstract}

\section{Introduction} 
As the prototypical unit of human activities, human-object interaction (HOI) plays an important role in activity understanding. Recently, image-based HOI learning~\cite{hico-det,ican,gpnn,idn,liu2022interactiveness,wu2022mining,decaug} achieves great progress. However, daily HOIs may need temporal cues to avoid ambiguity in detection, \eg, \textit{pick\_up-cup} and \textit{put\_down-cup}. Thus, many video HOI works~\cite{epic-kitchen,daly,cad++,ji2020action,trn,doh,SomethingElse,laptev2007retrieving,prest2012explicit} are proposed to advance spatio-temporal HOI (ST-HOI) learning.

\begin{figure}
    \centering
    \includegraphics[width=0.9\linewidth]{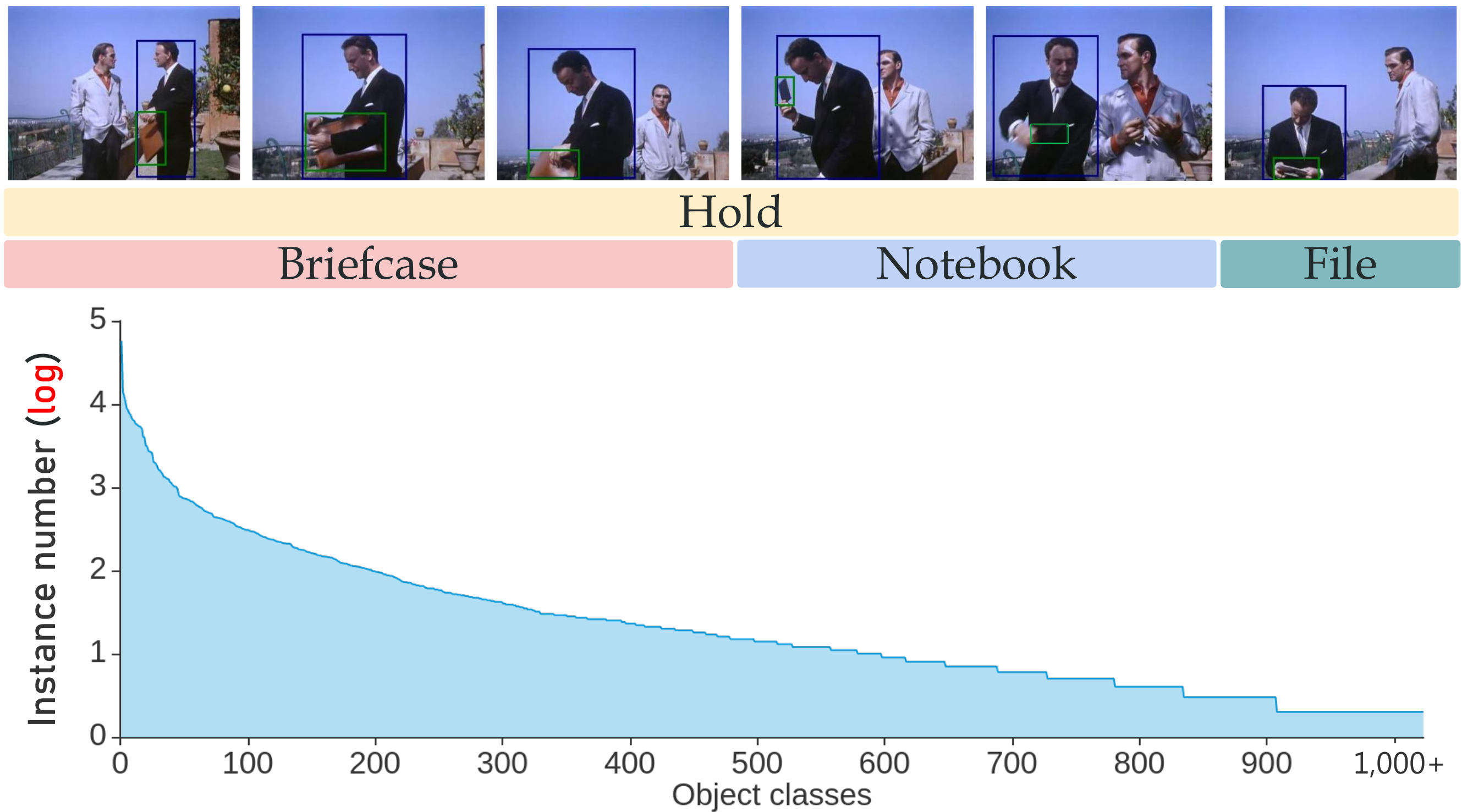}
    \vspace{-0.3cm}
    \caption{Object diversity of DIO. 
    In daily HOIs, we often interact with many objects.  
    DIO is rich in object classes and severely long-tailed, thus is more challenging.} 
    \label{fig:insight}
\vspace{-0.6cm}
\end{figure}

HOI data is often long-tailed, \ie, some HOIs are much more common than rare ones, which imports great challenges to vision systems~\cite{hico-det,AVA}. 
Generally, HOI is depicted as $\langle human,verb,object \rangle$ triplet. As the subjective is always human, thus rare HOIs are usually imported from rare \textit{verbs} or \textit{objectives}. 
For rare verbs, few-shot action learning has been studied in many works~\cite{hico-det,kato2018compositional,AVA}, but the rare objective problem is previously overlooked. 
In daily HOIs, we usually perform \textbf{limited} actions upon a variety of objects. For example, in Fig.~\ref{fig:insight}, a person \textit{holds} three different objects in a very short time.
But many video HOI datasets~\cite{charades,daly,ji2020action} contain few object classes, \eg, Charades~\cite{charades}, DALY~\cite{daly}, Action Genome~\cite{ji2020action} all have less than 50 classes (Tab.~\ref{tab:dataset_compare}). This makes vision systems built upon them lack the ability to localize various objects. 
For another, some video HOI datasets~\cite{epic-kitchen,SomethingElse,doh} including diverse objects are proposed recently, \eg, EPIC-Kitchens~\cite{epic-kitchen} contains 323 classes, Something-Else~\cite{SomethingElse} includes 18 K uncurated classes, 100DOH~\cite{doh} provides 110.1 K objects of unknown class. 
However, they all focus on \textit{hand-object} interactions and \textit{first-view} videos (100DOH~\cite{doh} also has third-view videos).
As \textbf{whole body-object} interaction detection from \textbf{third-view} videos matters to numerous applications (\eg, service robots, health-care), we prefer to study third-view body-object interactions, such as \textit{ride/sit on} (chair, horse, \etc), \textit{enter/exit} (train, bus, \etc).
In this work, we propose a large-scale third-view ST-HOI benchmark building upon AVA~\cite{AVA}: \textbf{D}iscovering \textbf{I}nteracted \textbf{O}bjectives (\textbf{DIO}). 
It contains 51 interactions and \textbf{1,000}+ object classes to afford HOI learning including many rare/unseen objects. And \textbf{290} K object boxes in \textbf{290} K $\langle human,verb,object \rangle$ frame-level triplets are provided.

Accordingly, an ST-HOI learning task and a corresponding integrated metric are proposed, requiring visual systems to track human actors and detect interactions while localizing interacted objects simultaneously. 
Notably, entity-relationship detection~\cite{vidvrd,hico-det,hakev2} often assumes that object locations can be obtained beforehand via detection~\cite{fasterrcnn}/tracking~\cite{milan2016mot16}. But this is difficult considering the diverse interacted object classes (Fig.~\ref{fig:insight}). 
In our task, object discovery is seen as \textit{important} and \textit{exploratory} as interaction detection. 
To this end, we split DIO to make its test set contain numerous \textit{rare} ST-HOI triplets composed of \textit{seen} interactions and \textit{rare/unseen} objects. 
Since few/zero-shot object detection~\cite{bansal2018zero} is still an open problem, current vision systems would struggle. For example, cutting-edge image/video detectors~\cite{fasterrcnn,video-detector} finetuned on our train set all achieve less than 20 AP. 
Hence, the exploration of \textbf{interacted object discovery} in DIO is extremely challenging but essential as the touchstone for the deep learning paradigm.

To tackle ST-HOI learning, we study how to track human actors, detect interactions and discover various objects with multiple points of view (Sec.~\ref{sec:experiment}) and propose a baseline system \textbf{H}ierarchical \textbf{P}robe \textbf{N}etwork (\textbf{HPN}).
It leverages a \textit{hierarchical-probe} policy to discover objects. 
Hierarchical-probe indicates that utilizing the spatio-temporal cues hierarchically, such as from \textit{local human parts}, to \textit{whole human body} and \textit{global context}, to discover possible interacted objects (Sec.~\ref{sec:method}).
In extensive experiments, HPN achieves impressive performance. 
However, DIO remains challenging and the three sub-tasks still have lots of room for improvement.
We believe DIO would inspire a new line of studies and pose new challenges and opportunities for the development of deeper activity understanding.

Our contributions are three-fold:
(1) We propose a large-scale third-view ST-HOI benchmark DIO, including 299 videos, 51 interactions, and 1,000+ objects. 
(2) A novel ST-HOI learning task and its metric are proposed to drive the studies on finer-grained activity parsing and understanding. 
(3) Accordingly, a baseline system, Hierarchical Probe Network (HPN), is proposed and achieves impressive performance on DIO.

\section{Related Works}
\label{sec:related-work}
\noindent{\bf Object Tracking.}
Object tracking is an active field and has two main branches, \ie, Single-Object Tracking (SOT)~\cite{siamban,fan2019lasot} and Multi-Object Tracking (MOT)~\cite{milan2016mot16,ristani2016performance,braso2020learning}.
Recently, tracking-by-detection (TBD)~\cite{kim2015multiple,sadeghian2017tracking} receives lots of attention and achieves state-of-the-art performance. 
As our main goal is object discovery, we directly adopt the cutting-edge DeepSORT~\cite{deepsort} and FairMOT~\cite{fairmot} as human trackers.

\noindent{\bf Human-Object Interaction (HOI).}
In terms of image-based HOI learning, both image-level~\cite{hico,kato2018compositional} and instance-level~\cite{hico-det,ican,tin,gkioxari2018detecting,idn,hakev1,fang2021dirv,liu2022highlighting,tinpami,djrn} methods achieve successes with the help of large-scale datasets~\cite{hico-det,hico,pastanet,hakev2}.  
As for HOI learning from third-view videos, recently many large-scale datasets~\cite{AVA,charades,daly,ji2020action,doh,vlog,activitynet} are released to promote this field, thus providing a data basis for us. 
They provide clip-level~\cite{activitynet,vlog,charades} or instance-level~\cite{AVA,ji2020action,daly} action labels, but few of them~\cite{vlog,charades,daly,ji2020action} afford diverse object classes. 
Though some datasets~\cite{SomethingElse,epic-kitchen} provide instance labels of diverse object classes, they usually concentrate on egocentric hand-object interaction understanding~\cite{xu2022learning}.
Relatively, we choose to focus on whole body-object interaction learning based on third-view videos and propose DIO featuring the discovery of diverse objects.
Recently, there are also many methods studying video-based visual relationship~\cite{vidvrd,liu2020beyond,tsai2019video} and HOI~\cite{trn,gpnn,strg,ma2018attend,baradel2018object,videotransformer}.

\noindent{\bf Object Detection.} 
Object detection~\cite{fasterrcnn,yolo} achieves huge success along with the development of deep learning and large-scale datasets~\cite{coco}, but they may struggle without enough training data.
Thus, some works~\cite{fsod,bansal2018zero} start to study few/zero-shot detection.
Moreover, as videos can provide temporal cues of moving objects, video object detection~\cite{video-detector} also receives attention recently.
Different from typical detection, some studies try to utilize context cues, such as human actor~\cite{kim2020tell,gkioxari2018detecting}, action recognition~\cite{yuan2017temporal,yang2019activity}, object relation~\cite{hu2018relation}, to advance object localization. 
Gkioxari~et~al.~\cite{gkioxari2018detecting} treat object localization as density estimation and use a Gaussian function to predict object location.
Kim~et~al.~\cite{kim2020tell} borrow cues from human pose and language prior to constructing a weakly-supervised detector.
In this work, we tackle object discovery via hierarchical perspectives, \ie, local human part, human, and global context. 

\section{Constructing DIO}
\label{sec:datset}
\noindent{\bf Data Collection.}
To support practical ST-HOI learning, we collect third-view videos from large-scale dataset AVA~\cite{AVA}. It contains 430 videos with spatio-temporal labels of 80 atomic actions (body motions and HOIs). As AVA includes complex HOIs in diverse scenes, it can bring great visual diversity to our benchmark.
We extract the HOI-related frames and the corresponding human boxes and action labels, thus the clips in DIO have uneven temporal durations. Notably, we only consider the non-human objectives in HOIs. 
Overall, based on the available train and validation (val) sets of AVA 2.2~\cite{AVA} (299 videos), we choose 74 hours of video including 51 HOIs (detailed in the supplementary).

\begin{figure*}[!ht]
	\begin{center}
        \begin{minipage}{.45\linewidth}
            \centerline{\includegraphics[width=\linewidth]{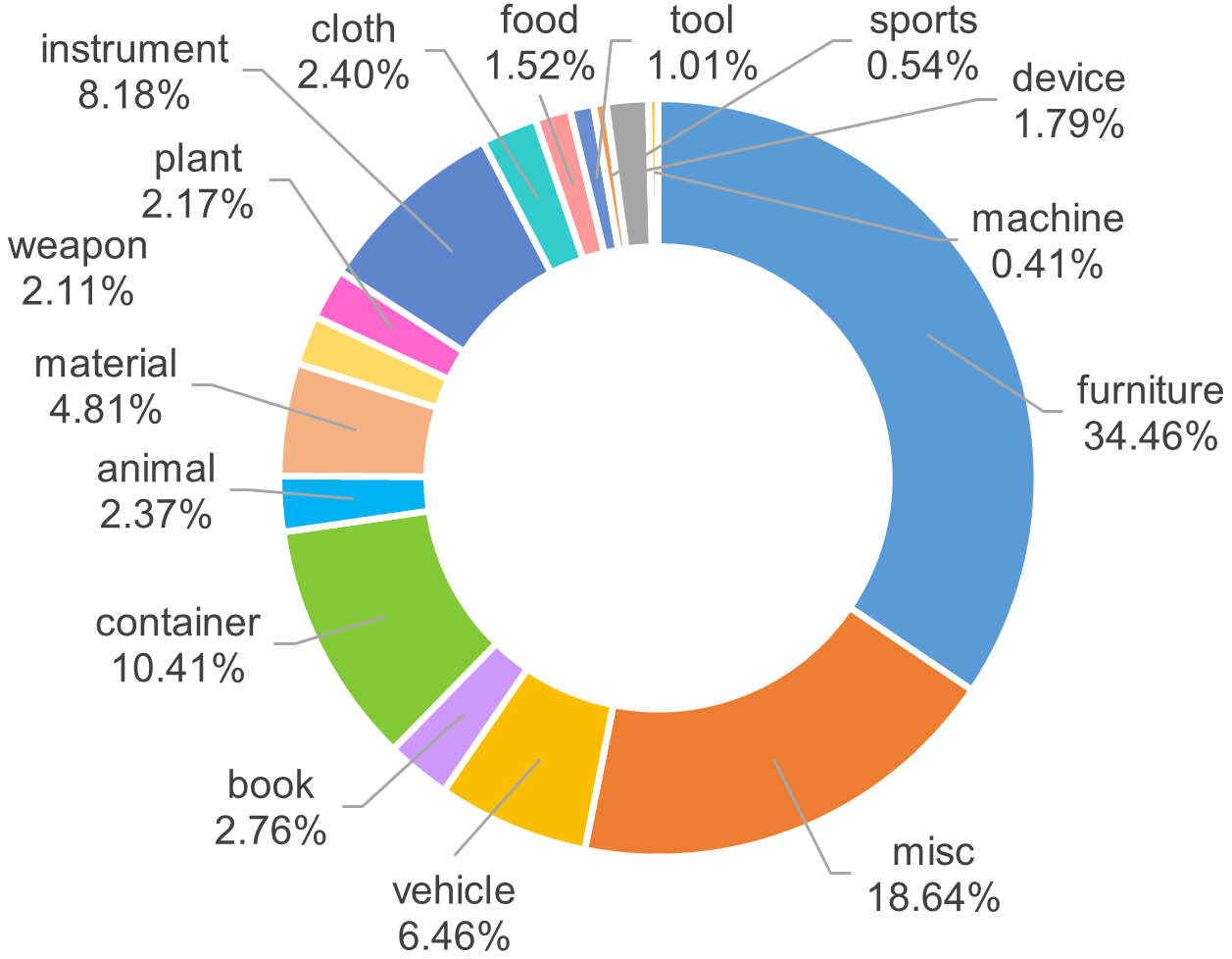}}
            \centerline{(a) Objects in DIO.}
        \end{minipage}
        \hfill
        \begin{minipage}{.275\linewidth}
            \centerline{\includegraphics[width=\linewidth]{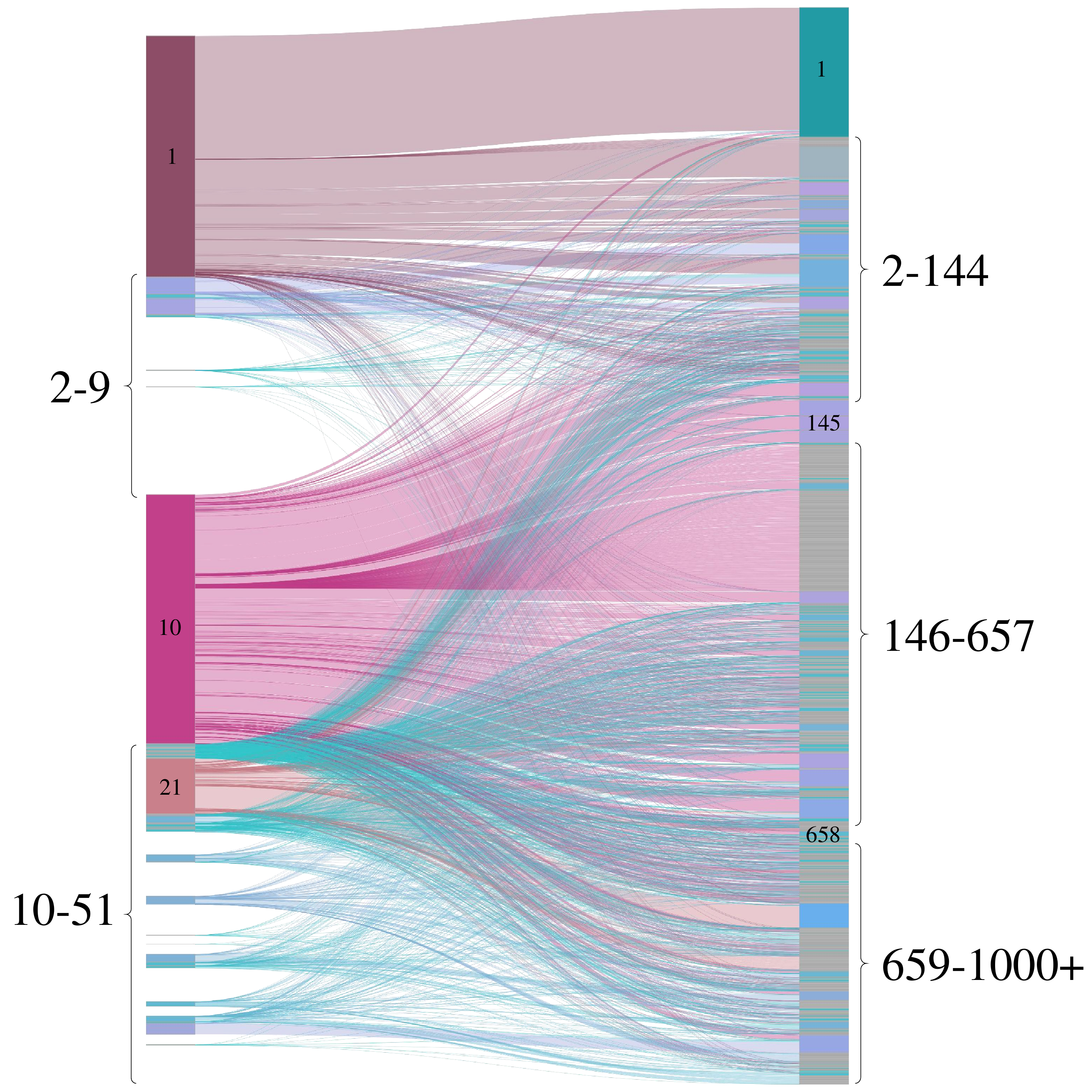}}
            \centerline{(b) Interaction-object co-occurrence.}
        \end{minipage}
        \hfill
        \begin{minipage}{.265\linewidth}
            \centerline{\includegraphics[width=\linewidth]{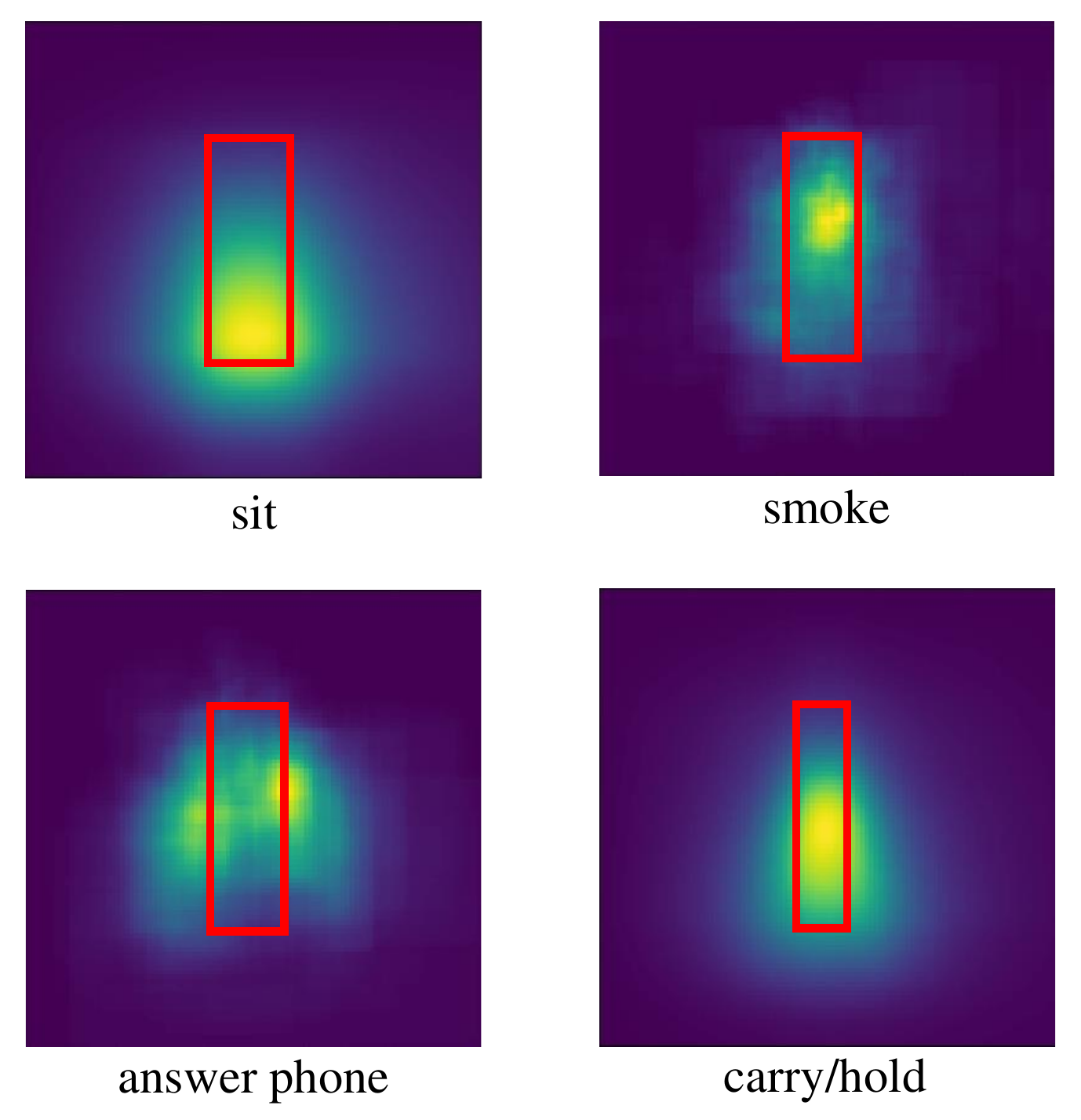}}
            \centerline{(c) Object location heatmap.} 
        \end{minipage}
	\end{center}
	\vspace{-0.5cm}
	\caption{DIO characteristics. (a) Objects in DIO. (b) Co-occurrence between interactions (left) and objects (right). (c) Object location heatmaps in the train set. Red boxes indicate the normalized human location.}
	\label{fig:characteristics}
	\vspace{-0.6cm}
\end{figure*}

\noindent{\bf Object Annotation.}
AVA provides labels with a stride of 1 second, so we add boxes and class labels for all interacted objects with the same stride. 
\textbf{First}, as a human can perform multi-interaction simultaneously, we set the annotating unit as a clip including one \textit{single} interaction. 
For example, a 30s clip including an actor \textit{holds-sth} (1-30s) and \textit{inspects-sth} (10-15s), will be divided into two sub-clips, \ie, a 30s sub-clip for \textit{holds-sth} and a 5s sub-clip for \textit{inspects-sth}. 
In brief, each sub-clip contains \textbf{one} verb and \textbf{one/several} class-agnostic interacted objects. 
Then, sub-clips are annotated separately, and each one is annotated by at least 3 annotators and checked by an expert to ensure quality.
\textbf{Second}, as AVA contains various scenarios and diverse objects, to better locate objects and avoid ambiguity, each annotator is given a whole sub-clip to draw boxes and classify them. 
In default, we use COCO~\cite{coco} 80 objects as a class pool. If annotators think an object is not in the pool, they are asked to input a suitable class according to their judgments. If an object cannot be recognized, they can choose the ``unknown'' option, but the ``unknown'' ratio is controlled by our annotation tool. 
Finally, we find that surprising 42.66\% of object instances are beyond our pool.
\textbf{Third}, after exhaustive annotation, we fix the input typos, exclude outliers via clustering, and combine similar items. 
After cleaning, \textbf{1,000}+ classes are extracted. We then conduct re-recognition for the frames including ``unknown'' objects. 
Finally, only 7,476 frames (2.57\%) still contain ``unknown'', due to the blur frames or too small objects. 
Fig.~\ref{fig:characteristics} (a) shows the object class composition according to WordNet~\cite{wordnet}. 
For more details about classes please refer to our supplementary.

\noindent{\bf New Challenge.}
Notably, the class labels are used for analysis instead of evaluation here, as current detectors~\cite{fasterrcnn,yolo,video-detector} cannot detect such an amount of classes. For example, a recent large-scale dataset FSOD~\cite{fsod} only includes \textbf{20.85}\% of our classes.  
And for practical applications, class-agnostic interacted object discovery is also essential for HOI learning, taking the situation in Fig.~\ref{fig:insight} as an example.
Hence, in our train set, 328 object classes only have less than 5 samples (boxes), and 98 classes are \textbf{unseen} in the testing. 
As few/zero-shot detection is beyond the scope of our work, we leave the opportunity of a more difficult setting including object classification to future work. 

\begin{table}[t]
\resizebox{\linewidth}{!}{
\setlength{\tabcolsep}{3.5pt}
\begin{tabular}{l cc cc cc cc cc}
\hline
 \multirow{2}{*}{Dataset}  & \multirow{2}{*}{Frames} & \multirow{2}{*}{Actions} & \multicolumn{2}{c}{Objects} & \multicolumn{2}{c}{HOI} &\multirow{2}{*}{View} &\multirow{2}{*}{Subjective}\\ 
 \cline{4-5}\cline{6-7}
 & & & class & instance & class & triplet & \\ 
\hline
\hline
Something-Something~\cite{goyal2017something} & 108K & 174 & - & - & 174 & - & first & hand\\
100DOH~\cite{doh} & 100K & 11 & - & 110.1K & 5 & 189.6K & first, third & hand\\
Something-Else~\cite{SomethingElse} & 8M & 174 & 18K*
& 10M & 174 & 6M & first & hand\\
EPIC-Kitchens~\cite{epic-kitchen}  & 266K & 125 & 323 & 454K & 125 & 243K & first & hand\\ 
CAD120++~\cite{cad++} & 65K & 10 & 13 & 64K & 2 & 60K & third & head, hand\\
VLOG~\cite{vlog}  & 114K & 9 & 30 & - & 9 & - & first, third & hand\\ 
\hline
AVA~\cite{AVA} & 387K & 80 & - & - & 51 & - & third & whole body\\  
Charades~\cite{charades} & 66K & 157 & 46 & 41K  & 157 & - & third & whole body\\
DALY~\cite{daly} & 11.9K & 10 & 43 & 11K & 10 & 11K & third & whole body\\
Action Genome~\cite{ji2020action} & 234K & 157 & 35 & 476K & 25 & 1.72M & third & whole body\\
\hline
DIO & 126K  & 51 & \textbf{1,000}+ & 290K & 51 & 290K & third & whole body\\
\hline
\end{tabular}}
\vspace{-0.3cm}
\caption{Dataset comparison. 
Frames with labels are included. Instances/triplets are in \textit{frame-level}.
For clip-level annotations, we count a clip as a frame.
For Action Genome~\cite{ji2020action}, we count the HOIs and exclude the other relations. 
\textit{18 K*}: object class labels of~\cite{SomethingElse} are uncurated.}
\label{tab:dataset_compare}
\vspace{-0.5cm}
\end{table}

\noindent{\bf ST-HOI Tracklet.}
Since one human may interact with multi-object via multi-interaction simultaneously, we largely decrease the complexity via the above sub-clip segmentation to decouple interactions.
Next, to generate the ST-HOI labels, we further consider the objects in each sub-clip (one interaction of a person).
\textbf{First}, if there is only one object in a sub-clip, we use its locations as the labels.
If there are multiple objects, we record all of their boxes and manually link their boxes as multiple object tracklets.
Besides, to lower the level of discovery difficulty and simplify the evaluation, if a model can accurately discover \textit{one} of the multiple GT objects, object discovery would be seen successfully.
That said, the largest overlap between the predicted object and GT objects is calculated.
\textbf{Second}, considering the costly annotation and the demand of class-agnostic object discovery, except the situation of the multi-GT object, we do not distinguish object instances in videos with \textit{object ID} like MOT~\cite{milan2016mot16}.
\textbf{Third}, each sub-clip is seen as a \textit{ST-HOI tracklet}, whose label records \textit{a human actor tracklet, an interaction, a/several class-agnostic object tracklets}.

\noindent{\bf Dataset Statistics}
DIO includes 290 K HOI triplets, and 290 K object boxes of 1,000+ classes (Tab.~\ref{tab:dataset_compare}). 
Compared to previous datasets, it shows an obvious demand for rare/unseen object discovery.
Some characteristics are shown in Fig.~\ref{fig:characteristics}.
We split the data into train and test sets with a ratio of 5:1 (videos).
For more please refer to the supplementary.

\subsection{Task and Metric}
\label{sec:task-metric}
\begin{figure}
    \centering
    \includegraphics[width=0.95\linewidth]{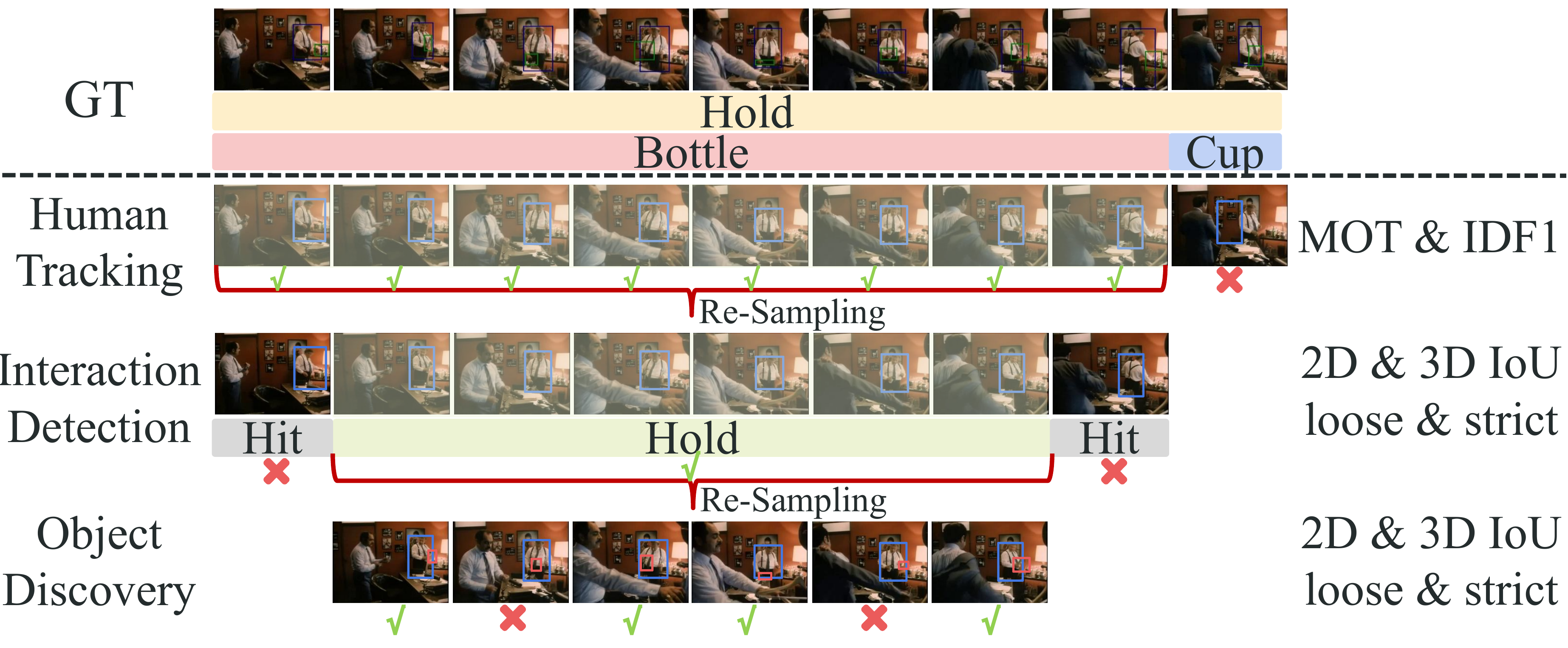}
    \vspace{-0.3cm}
    \caption{Task and metric settings. 
    We adopt MOT~\cite{milan2016mot16}, AVA~\cite{AVA}, SOT~\cite{kristan2018sixth} and a step-wise sampling strategy.
    }
    \label{fig:task}
\vspace{-0.6cm}
\end{figure}

Our task expects vision systems to detect ST-HOI triplets from videos. For a true positive, the \textbf{three elements} of ST-HOI, human tracking (location, id), interaction detection (location, class), and object discovery (location) all have to be accurate (Fig.~\ref{fig:task}).
For human tracking, our task is similar to MOT~\cite{milan2016mot16} but only expects the interacting actors instead of all persons.
For interaction detection, we follow the setting of AVA~\cite{AVA}, \ie, seeing it as a detection problem to detect the actors performing defined interactions.
For objective discovery, since numerous classes make conventional/few/zero-shot object detectors~\cite{bansal2018zero,fsod,fasterrcnn} struggle, we set this task as class-agnostic object discovery following SOT setting~\cite{kristan2018sixth}.
Accordingly, vision systems need to first localize humans accurately as the \textit{basis} for the subsequent two sub-tasks. Moreover, the quality of interaction detection also affects object discovery. Hence, three sub-tasks are \textbf{coupled} and make the evaluation difficult. 
To tackle this problem, we propose an \textbf{evaluation via step-wise sampling} strategy to gradually evaluate three sub-tasks (Fig.~\ref{fig:task}). 
Next, we detail the metrics and evaluations of three sub-tasks.

\noindent{\bf Human Tracking.}
Given human tracklets, we adopt the widely-used MOTA~\cite{milan2016mot16} and IDF1~\cite{ristani2016performance} to measure human detection and re-identification (Re-ID), which are fundamental for the subsequent two sub-tasks. 
Notably, a model may first track all persons and then use interaction prediction to \textit{exclude} the persons who do not perform HOIs.
A true positive actor tracklet should contain both accurate boxes (IoU$>$0.5 regarding GT boxes) and ID. As the annotation density of AVA is one frame per second, we evaluate two metrics according to the tracking results on the \textit{first frame of each second} and report the mean values of all videos.
According to the step-wise sampling, the \textbf{true positive} frames of actor tracking (Fig.~\ref{fig:task}, green checks) are sent to interaction detection evaluation, the rest of \textbf{false positive} frames (red crosses) are seen as false positives for two subsequent sub-tasks because the inaccuracy of actor tracking would introduce unreliable interaction and object predictions, \eg, nonexistent humans or inaccurate boxes. 

\noindent{\bf Interaction Detection.}
After measuring actor tracking, we next evaluate interaction detection, \ie, which actors are performing the defined interactions, when, and where they are performing. This is equal to instance-level action detection like AVA~\cite{AVA}.
All predictions are in the format of ST-HOI tracklet, \ie, successive frames of an actor including only one predicted continuous interaction. 
For all predictions, we adopt a threshold $\alpha$ to mask the interaction scores under $\alpha$ as zero.
That is, if an ST-HOI tracklet for the $i$-th interaction $a_i$ contains a frame with a score under $\alpha$, this frame would be excluded and this tracklet would be \textit{divided} into two ST-HOI tracklets.
We adopt two metrics: \textbf{2D} and \textbf{3D IoU}.
For 2D IoU, we calculate the 2D IoU between the predicted and GT boxes of each frame in an ST-HOI tracklet and get the average IoU of the whole tracklet.
For 3D IoU, we link the successive human boxes as a tube and measure the 3D IoU between predicted and GT polyhedrons within one second. The mean 3D IoU of a tracklet is also calculated.
A tracklet would be seen as a true positive if its mean 2D/3D IoU (mIoU) is higher than 0.2, following AVA~\cite{AVA}.
We also set \textbf{two criteria} here.
A \textbf{loose} one only measures mIoU on the true positive frames of actor tracking, thus a separate evaluation for interaction detection can be conducted.
Moreover, a \textbf{strict} one considers all frames, \ie, false positive frames of actor tracking (Fig.~\ref{fig:task}, red crosses) are seen as \textit{false positives} for interaction detection too. 

\noindent{\bf Interacted Object Discovery.}
An ST-HOI tracklet contains a/several class-agnostic GT object tracklets.
For simplicity, we treat object discovery as a SOT problem and follow its metric~\cite{kristan2018sixth}. We adopt the average overlap metric, \ie, mIoU of all frames in an ST-HOI tracklet. 
If there are multiple GT object tracklets, we calculate all IoUs between the predicted tracklet and GT tracklets and save the largest IoU as the result. 
\textbf{Two criteria} are also used: 
(1) In the \textbf{loose} one, false positive frames from both actor tracking and interaction detection (Fig.~\ref{fig:task}, red crosses) are all discarded and not considered in the evaluation. It also provides a separate evaluation for interacted object discovery.
(2) On the contrary, the \textbf{strict} one directly treats all false positive frames from the former two sub-tasks (red crosses) as \textit{false positives} of object discovery, and their mIoUs are set to zero. The rest of the frames (green checks) are evaluated normally.

\section{Method}
\label{sec:method}
In this section, we describe the pipeline of HPN (Fig.~\ref{fig:HPN}). 
For clarity, the description unit hereinafter is \textit{one human tracklet} including one tracked person. 

\subsection{Human Tracking}
Given a clip $C$, we first adopt FairMOT~\cite{fairmot} or DeepSORT~\cite{deepsort} to obtain the human tracklets: $T_h = \mathcal{F}_{track}(C)$.
A human tracklet is represented as $T_h=\{I^k_h\}^n_{k=1}$ ($n$ seconds).
For $T_h$, we operate interaction detection next to divide it into ST-HOI tracklets including one interaction each.

\subsection{Interaction Detection}
\label{sec:action-detection}
Here, we adopt SlowFast~\cite{feichtenhofer2019slowfast}.
As we re-split AVA, our test set contains some clips from the AVA train set, we use Kinetics~\cite{kinetics} pre-trained backbone and finetune it on our train set to avoid \textit{pollution}.
Given a human tracklet $T_h=\{I^k_h\}^n_{k=1}$, we use 8 frames/s for slow branch and 32 frames/s for fast branch. We input slow ([$8, 2048, 7^2$]) and fast ([$32, 256, 7^2$]) RoI-Align features into the pooling layers and concatenate them as a $2,304$ sized tensor, then feed it into a fully-connected (FC) layer to obtain the action score $S^{k}_{a} = \mathcal{F}_{action}(I^k_h)$, where $S^{k}_{a}=\{S^{ki}_{a}\}^{51}_{i=1}$, $k,i$ indicate the indexes of second and interaction class.
Interaction probabilities $p^{k}_a=Sigmoids(S^{k}_a)$.
We use binary cross entropy losses for 51 interaction classes $L^{k}_a=(\sum^{51}_{i=1} L^{ki}_a)/51$, the whole action loss of $T_h$ is $L_a=\sum^{n}_{k=1} L^{k}_a$.
The scores under threshold $\alpha$ will be masked to zero. If all scores of $T_h$ are masked, we would delete $T_h$.
Then, we divide $T_h$ into ST-HOI tracklets including successive frames of a single interaction. As for the masking operation, one interaction may have multiple ST-HOI tracklets. 
For clarity, in $T_h$, we use $T^{i}_{HOI}$ to represent the ST-HOI tracklet of $a_i$ and omit the possible multi-tracklet. 

\begin{figure}
    \centering
    \includegraphics[width=0.97\linewidth]{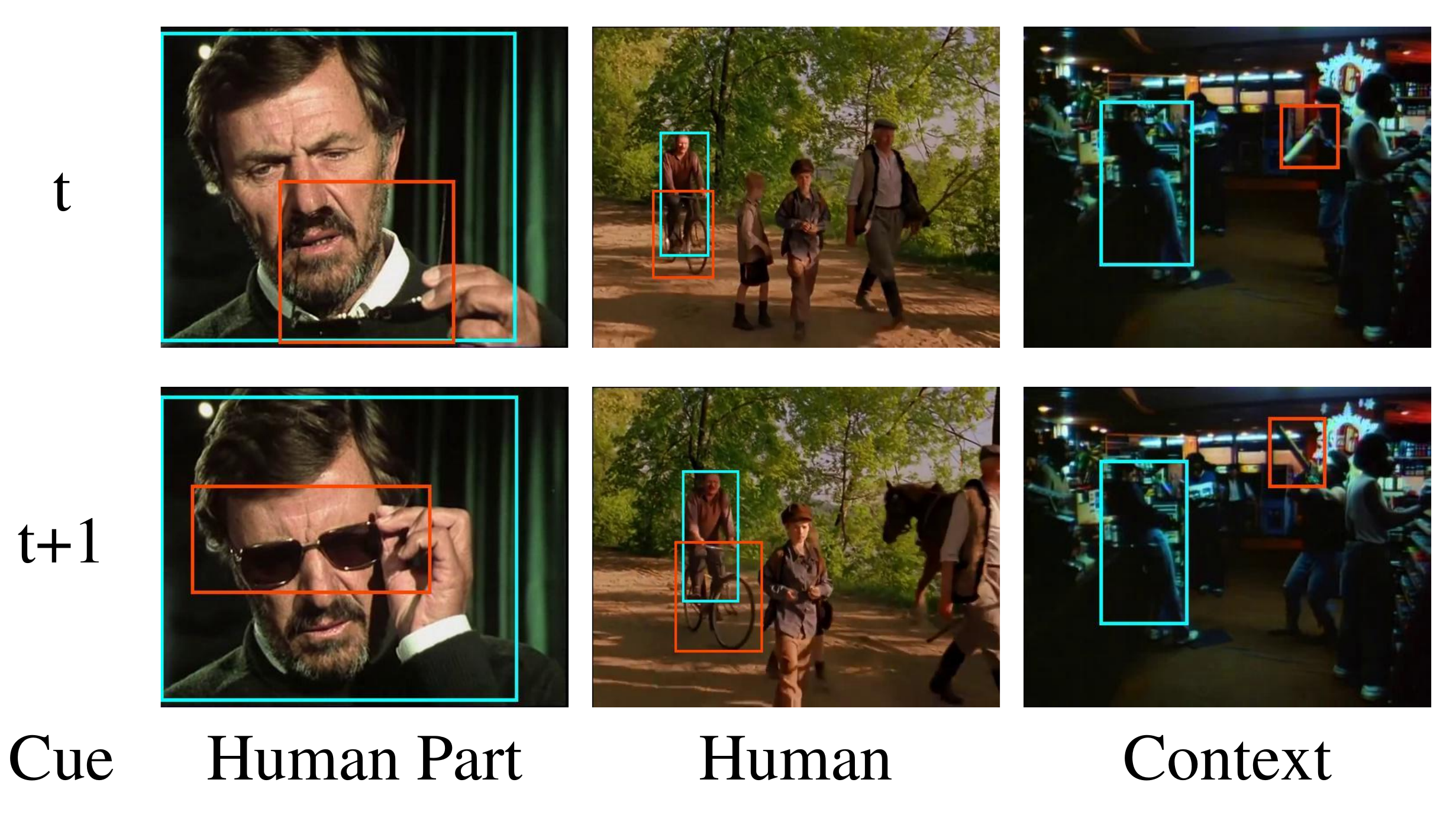}
    \vspace{-0.3cm}
    \caption{Three scenarios for interacted object discovery.} 
    \label{fig:three-scenarios}
\vspace{-0.5cm}
\end{figure}

\begin{figure*}
	\begin{center}
        \centerline{\includegraphics[width=0.9\linewidth]{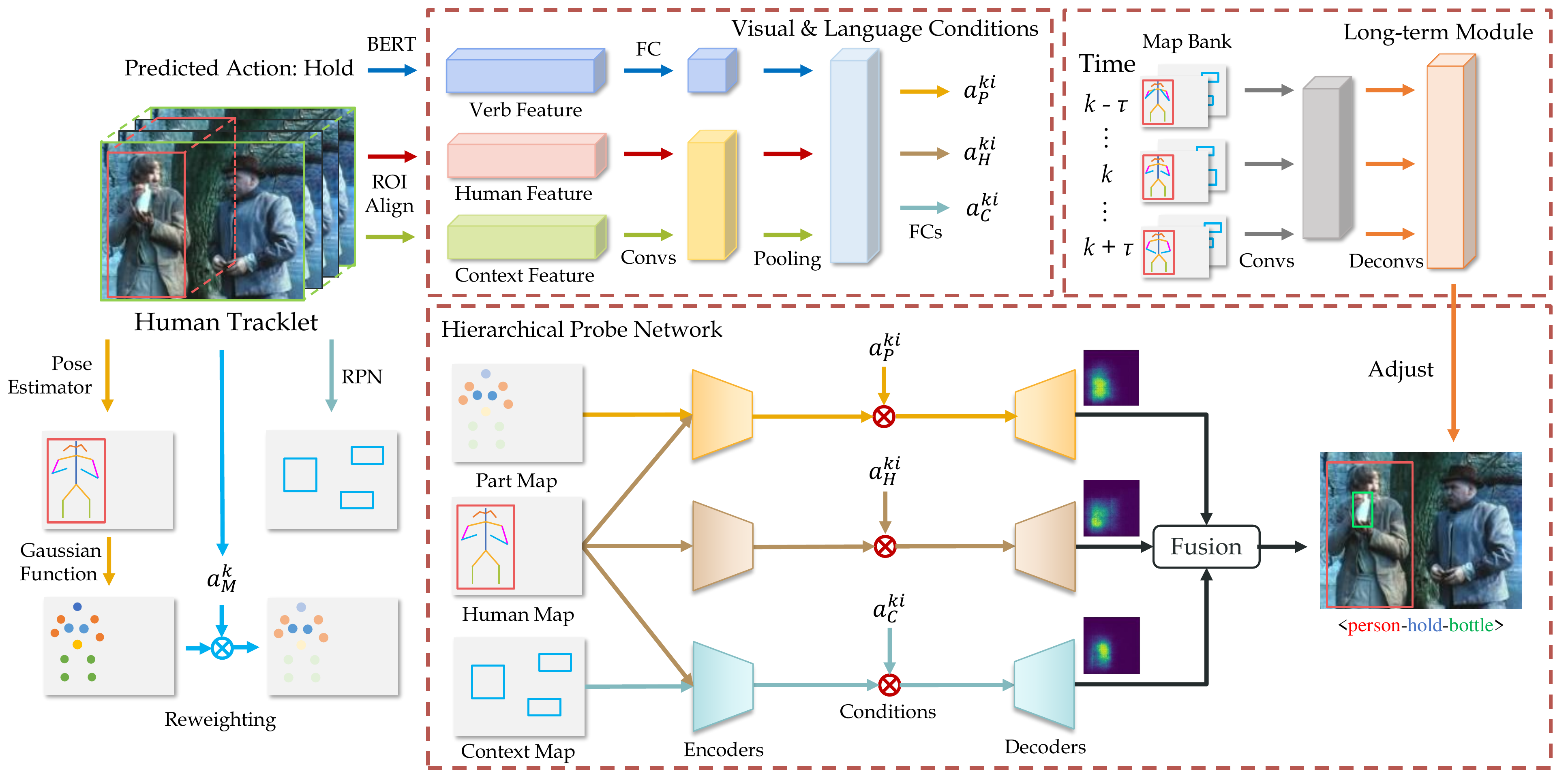}}
	\end{center}
	\vspace{-1cm}
	\caption{The overview of baseline HPN. Our hierarchical-probe policy discovers objects from 3 perspectives, \ie, local human parts, human, and global context. We leverage the spatio-temporal configuration (STC) to construct three STC maps.} 
    \label{fig:HPN} 
\vspace{-0.6cm}
\end{figure*}

\subsection{Interacted Object Discovery}
\label{sec:object}
\noindent{\bf Limitations of Current Methods.}
It is difficult to detect objects via conventional detectors~\cite{fasterrcnn,yolo} since the diverse/unseen objects. And zero/few-shot object detection~\cite{fsod,bansal2018zero} also does not work well.
For example, after finetuning the image-based RPN~\cite{fasterrcnn} and video-based detector~\cite{video-detector} on our train set with objectiveness detection, their performances are only \textbf{12.47} and \textbf{12.25} AP.
Thus, it is essential to borrow video cues to help object discovery. 

\noindent{\bf Typical ST-HOI Scenarios.}
To get some inspiration, let us check three typical scenarios (Fig.~\ref{fig:three-scenarios}). 
For the left $\langle human, hold, glasses \rangle$ including small glasses, 
we can find some cues from the moving hand, \ie, \textit{human part}.
For the middle $\langle human, ride, bicycle \rangle$, as the bicycle is moving closely with the human, the \textit{human body} can act as a cue.
For the $\langle human, listen\_to, instrument \rangle$, besides the human, the context (objects around human) may help us.

\noindent{\bf Hierarchical-Probe.} 
Hence, we propose a hierarchical-probe policy to leverage three perspectives as the \textbf{starting points} of discovery. \textit{Hierarchical} indicates that the interested region varies from \textbf{local human part}~\cite{pastanet,partstate} to \textbf{whole human}, then to \textbf{global context}.
From three perspectives, we utilize visual and semantic video contents as the \textbf{conditions} to guide HPN to ``search'' objects. 
In an ST-HOI tracklet $T^{i}_{HOI}=\{I^k\}^m_{k=1}$ ($m$ seconds), for its $k$-th second frames $I^k$, the visual conditions are $f^{k}_h$ (human boxes) and $f^k_c$ (whole frames), which are the features after the temporal pooling ([$2304, 7^2$]) of SlowFast; the semantic condition is $f^i_v$ (Bert~\cite{bert} vector of $a_i$):
\begin{equation}
    \begin{aligned}
    \label{eq:conditions}
        f^{k}_h, f^k_c = \mathcal{F}_{slowfast}(I^k),
        f^i_v = \mathcal{F}_{bert}(a_i).
    \end{aligned}
\end{equation}
Next, we construct the ``search'' starting points as three \textbf{spatio-temporal configuration (STC)} maps (Fig.~\ref{fig:HPN}).

\noindent{\bf (1)}
For \textbf{human part}, we first use pose estimation~\cite{fang2017rmpe,alphapose,li2019crowdpose,li2021human} to obtain 18 body key points. 
For each keypoint, we prepare a [56, 56] grey-scale map including a two-dimensional Gaussian distribution centering at the keypoint with $\sigma_1=\sigma_2=3$ pixels, to indicate part location. 
To distinguish the parts, we concatenate $f^{k}_h, f^i_v$ and input them into an MLP-Sigmoids to generate 18 attention values $a^{k}_M$ to reweight 18 keypoint maps.
Then, we concatenate reweighted maps with human skeleton and box maps~\cite{tin}, which are all [56, 56] to represent the locations of parts relative to the whole body.
This [20, 56, 56] tensor is used as \textit{human part STC map} $M^{k}_P$.

\noindent{\bf (2)}
For \textbf{whole human body}, we use human skeleton and box maps ([2, 56, 56]) as the \textit{human STC map} $M^{k}_H$.

\noindent{\bf (3)}
For \textbf{context}, we finetune an RPN~\cite{fasterrcnn} on our train set and detect the objectness in all frames. The predicted proposals after NMS are selected as the important objects in context. Thus we can construct object maps~\cite{tin} with these proposals. Next, we concatenate them with human box and skeleton maps ([2+$N_p$, 56, 56]) as the \textit{context STC map} $M^{k}_C$, where $N_p$ is the number of proposals.

To leverage the hierarchical cues, we adopt a multi-branch structure (Fig.~\ref{fig:HPN}) consisting of three branches. For detailed structure please refer to our supplementary.

\noindent{\bf Part Branch.}
Corresponding to the \textit{human part STC map} $M^{k}_P$, we use a convolutional encoder to embed it and obtain the compressed $\hat{M^{k}_P}$. And the condition $\langle f^{k}_h, f^i_v \rangle$ is also compressed into $f^{ki}_{hv}$. 
Then, we use a convolutional-Sigmoids module to transfer $f^{ki}_{hv}$ to attention tensor $a^{ki}_P$ which has the same channel numbers with $M^{k}_P$. The adjustment upon the part cue based on the condition is implemented as element-wise multiplication:
\begin{equation}
    \begin{aligned}
    \label{eq:compress_1}
        \hat{M^{ki}_P} = \hat{M^{k}_P} \otimes a^{ki}_P.
    \end{aligned}
\end{equation}
The superscript $ki$ means the $k$-th second and $a_i$ in an ST-HOI tracklet, and each predicted tracklet contains one person, one interaction and one class-agnostic interact object $b^{ki}_O$.
Next, we adopt a convolutional decoder to decode $\hat{M^{ki}_P}$ to $f^{ki}_P$ which is the same size as video frames. 
At last, we use pixel-level Sigmoids to translate $f^{ki}_P$ into a heatmap which can illustrate the object location: $h^{ki}_P = Sigmoids(f^{ki}_P)$.
For each GT object box, a chosen GT heatmap $h^{ki}_{GT}$ is constructed by generating a 2-dimensional Gaussian distribution centering at the GT box center with $\sigma_1, \sigma_2$ as half of the box width and height.
The pixel-wise Sigmoid cross entropy (BCE) loss $L^{ki}_P$ is employed for $h^{ki}_P$. 

\noindent{\bf Human and Context Branches.}
Their structures are similar to the part branch. 
First, we compress $M^{k}_H, M^{k}_C$ into $\hat{M^{k}_H}, \hat{M^{k}_C}$ with two convolutional encoders.
Then the conditions $\langle f^{k}_h, f^i_v \rangle$ for human branch and $\langle f^{k}_h, f^{k}_c, f^i_v \rangle$ for context branch are compressed into $f^{ki}_{hv}$ and $f^{ki}_{hcv}$ respectively. 
We use two convolutional-Sigmoids modules to transfer $f^{ki}_{hv}$ to $a^{ki}_P, a^{ki}_H$, and $f^{ki}_{hcv}$ to $a^{ki}_C$.
The adjustments are similar:
\begin{equation}
    \begin{aligned}
    \label{eq:compress_2}
        \hat{M^{ki}_H} = \hat{M^{k}_H} \otimes a^{ki}_H,\\
        \hat{M^{ki}_C} = \hat{M^{k}_C} \otimes a^{ki}_C.
    \end{aligned}
\end{equation}
Two decoders are then used to decode $\hat{M^{ki}_H}, \hat{M^{ki}_C}$ to $f^{ki}_H, f^{ki}_C$. The two output heatmaps are obtained via $h^{ki}_H = Sigmoids(f^{ki}_H), h^{ki}_C = Sigmoids(f^{ki}_C)$.
And the corresponding pixel-wise BCE losses are $L^{ki}_H, L^{ki}_C$. 

\noindent{\bf Object Size Classification.} 
As objects with different sizes have different localization difficulties, \eg, small objects are usually harder to discover (proven in Sec.~\ref{sec:experiment}).
To enable HPN to distinguish object size and take corresponding heatmap filtering strategies, we further adopt a simple object size classifier.
First, for objects in train and val sets, we calculate the areas of GT human $a_h$ and object $a_o$ and calculate the ratio between them: $r_{o|h}=a_o/a_h$.
Then, $r_{o|h}$ is used to separate the GT objects into three size classes: small~($r_{o|h} \le 0.3$), medium~($0.3 < r_{o|h} \le 1$), large~($r_{o|h} > 1$). 
Finally, we introduce a two-layer MLP classifier with Softmax, its input is the concatenation of $f_{c}^k$ and $f_{h}^k$. For the object of the $k$-th second and $i$-th interaction, it would predict its size class. A cross-entropy loss $L^{ki}_{size}$ is used to train it.
In testing, the classified size would decide the heatmap filtering threshold (Sec.~\ref{sec:implementation}).

\noindent{\bf Perspective Fusion.}
For heatmaps $h^{ki}_{P/H/C}$, after normalizing and filtering with a threshold, we can generate the tightest bounding boxes for the binary masks as the predicted locations.
As the three branches are complementary, we fuse their predictions with three strategies:
\textbf{(1) Equal Fusion}: directly fusing $h^{ki}_P, h^{ki}_H, h^{ki}_C$ with $h^{ki}=1/3(h^{ki}_P + h^{ki}_H + h^{ki}_C)$ to get box $b^{ki}_O$.  
\textbf{(2) Box Selection}: obtaining boxes from heatmaps, \ie, $b^{ki}_{PO}, b^{ki}_{HO}, b^{ki}_{CO}$, and building a val set by randomly choosing 50 videos from train set to evaluate them. For $a_i$, we select a branch prediction that performed best on val set.
\textbf{(3) Dynamic Fusion}: concatenating $\hat{M^{k}_P}, \hat{M^{k}_H}, \hat{M^{k}_C}$ and inputting them into an FC-Softmax to generate dynamic fusion factors $\beta_P, \beta_H, \beta_C$, then $h^{ki}=\beta_Ph^{ki}_P + \beta_Hh^{ki}_H + \beta_Ch^{ki}_C$. 
On val set, dynamic fusion performs best and is chosen as our final setting.

\noindent{\bf Long-Term Information.} 
Besides the short-term cues of one second utilized above, long-term cues are also proven effective in video understanding~\cite{lfb}. 
Thus, we adopt a module to capture long-term cues and generate a heatmap to complement the short-term prediction. 
For the $k$-th second, we construct a \textit{STC map flow} belonging to time range $[k-\tau, k+\tau]$ by concatenating the STC (skeleton, human box, proposals) maps of $[k-\tau, k+\tau]$, where $\tau$ is the time offset and set as 2 in practice. Insufficient seconds will be filled with empty maps as placeholders.
For the STC flow ($[N, 2+N_p, 2\tau+1, 56, 56]$), we use three 3D convolution layers to aggregate the local ST information, a non-local layer to pass the information between local ST ranges, a temporal pooling layer to associate in the time dimension, and five 2D deconvolution layers to generate the heatmap.
The heatmaps from short-term and long-term modules are fused with weights $\epsilon$ and $1-\epsilon$. 
Finally, we operate filtering upon the fused heatmap to obtain the final predicted object box. 

\noindent{\bf Losses.}
The object discovery loss for $T^{i}_{HOI}=\{I^k\}^m_{k=1}$ is $L^{i}_o=\sum^m_{k=1} (L^{ki}_P + L^{ki}_H + L^{ki}_C + L^{ki}_{size})$. For a human tracklet $T_h$, the object discovery loss is $L_o = (\sum^{51}_{i=1}L^{i}_o)/51$. 
The total loss of $T_h$ is $L=L_a+L_o$ ($L_a$ in Sec.~\ref{sec:action-detection}).

\section{Experiments}
\label{sec:experiment}

\subsection{Settings}
We report the results following Sec.~\ref{sec:task-metric}.
For human tracking, we calculate the MOTA~\cite{milan2016mot16} and IDF1~\cite{ristani2016performance}. After step-wise sampling, the 2D/3D mIoU under strict/loose criteria is reported for action detection and object discovery.

\subsection{Baselines} 
\label{sec:baseline}
To tackle challenging interacted object discovery, we look back to previous methods, analyze their characteristics and propose several baselines.
We present the baselines in descending order of the reliance on object detection.
As our main goal is to explore the interacted object discovery, for a fair comparison, all baselines adopt the same human trackers, interaction classifier, and backbone with HPN.
More details are described in the supplementary.

\noindent{\bf Proposal-based Baselines}:
With the RPN~\cite{fasterrcnn} finetuned on the train set, we obtain the proposals of each frame in an ST-HOI tracklet. 
Here, four proposal-based baselines are adopted:
(1) \textbf{Proposal Selection (PS)}~\cite{tan2019learning} selects the proposal generating the highest interaction classification score.
(2) \textbf{Proposal+Tracking-By-Detection (PTBD)} uses TBD~\cite{zhang2008global} to match proposals into tracklets and chooses the one generating the highest mean interaction score.
(3) \textbf{Proposal+Single Object Tracking (PSOT)} tracks each proposal via SOT~\cite{siamrpn} and chooses the one generating the highest mean interaction score.
(4) \textbf{Proposal Adjustment (PA)}~\cite{chen2017learning} chooses a proposal with the highest objectness confidence and learns the offset between it and the GT box.

\noindent{\bf Human-to-Object (HTO)~\cite{gkioxari2018detecting}}
directly adjusts the human box to the object box via a learned offset. 

\noindent{\bf Density Estimation (DE)~\cite{gkioxari2018detecting}} 
sees object discovery as density estimation to estimate the object location likelihood and multiplies this likelihood with the interaction score.

\noindent{\bf Shortest Distance (SD)~\cite{doh}}
learns an offset from human to object similar to HTO and chooses a proposal having the shortest center distance with the HTO estimated box.

\noindent{\bf Heatmap (HM)~\cite{kim2020tell}}
directly inputs the \textit{condition} feature, skeleton map, and verb vector to estimate the heatmap.

\noindent{\bf Box Regression (BR)}
directly regresses the object box. 

\begin{table*}[t]
\centering
\resizebox{0.85\linewidth}{!}{%
\begin{tabular}{@{\extracolsep{4pt}}c c cc cc cc cc cc cc@{}}
\hline
 \multirow{3}{*}{Human Tracking} & \multirow{3}{*}{Methods} & \multicolumn{2}{c}{Human Tracking} & \multicolumn{4}{c}{Interaction Detection (mAP(\%))} & \multicolumn{4}{c}{Object Discovery (mIoU(\%))} \\ 
 \cline{3-4}\cline{5-8}\cline{9-12}
  & & \multirow{2}{*}{MOTA(\%)} & \multirow{2}{*}{IDF1(\%)} & \multicolumn{2}{c}{Strict} & \multicolumn{2}{c}{Loose} & \multicolumn{2}{c}{Strict} & \multicolumn{2}{c}{Loose} & \\ 
  \cline{5-6}\cline{7-8}\cline{9-10}\cline{11-12}
  & & & & 2D & 3D &  2D & 3D &  2D & 3D &  2D & 3D \\
  \hline
  \multirow{15}{*}{DeepSORT} 
  & Proposal Selection (PS) & 15.9
  & 44.1 & 9.5 & 5.5 & 10.2 & 6.0 & 1.1 & 0.9 & 14.0 & 17.3 \\
  & Proposal + TBD (PTBD) & 15.9 & 44.1 & 9.5 & 5.4 & 10.2 & 6.0 & 1.1 & 0.9 & 14.6 & 18.0  \\
  & Proposal + SOT (PSOT) & 15.9 & 44.1 & 9.3 & 5.4 & 10.1 & 6.2 & 1.1 & 0.9 & 13.9 & 17.1 \\
  & Proposal Adjustment (PA) & 15.9 & 44.1 & 9.3 & 5.4 & 10.0 & 6.0 & 1.3 & 1.0 & 16.7 & 20.5 \\
  & Human-to-Object (HTO) & 15.9 & 44.1 & 8.9 & 5.6 & 9.3 & 6.3 & 1.5 & 1.2 & 19.2 & 23.7 \\  
  & Density Estimation (DE) & \textbf{20.0} & 
  \textbf{44.6} & \textbf{11.9} & \textbf{6.8} & \textbf{12.3} & \textbf{7.5} & \textbf{2.5} & \textbf{2.1} & 22.1 & 26.4 \\
  & Shortest Distance (SD) & 15.9 & 44.1 & 9.4 & 5.4 & 10.1 & 5.9 & 1.4 & 1.1 & 18.0 & 22.3 \\ 
  & HeatMap (HM) & 15.9 & 44.1 & 8.3 & 5.1 & 9.0 & 5.5 & 1.4 & 1.1 & 18.3 & 22.6 \\
  & Box Regression (BR) & 15.9 & 44.1 & 8.7 & 5.4 & 9.3 & 5.8 & 0.9 & 0.7 & 12.2 & 14.9 \\ 
  \cline{2-12}
  & HPN-P & 15.9 & 44.1 & 8.4 & 4.7 & 8.8 & 5.2 & 1.4 & 1.1 & 20.6 & 25.4 \\
  & HPN-H & 15.9 & 44.1 & 8.2 & 4.7 & 8.6 & 5.2 & 1.4 & 1.1 & 19.7 & 24.2 \\
  & HPN-C & 15.9 & 44.1 & 9.0 & 5.2 & 9.5 & 5.8 & 1.6 & 1.3 & 22.2 & 27.8 \\
  & \textbf{HPN} & 17.3 & 44.2 & 9.7 & 5.4 & 10.0 & 6.1 & 1.6 & 1.3 & \textbf{24.6} & \textbf{30.4} \\
  \hline\hline
\multirow{15}{*}{FairMOT} 
  & Proposal Selection (PS)& 3.9 & 45.5 & 3.6 & 2.1 & 3.8 & 2.6 & 2.5 & 2.0 & 15.1 & 18.5 \\
  & Proposal + TBD (PTBD) & 3.9 & 45.5 & 3.6 & 2.1 & 3.7 & 2.6 & 2.6 & 2.1 & 15.6 & 19.1 \\
  & Proposal + SOT (PSOT) & 3.9 & 45.5 & 3.6 & 2.1 & 3.8 & 2.5 & 2.5 & 2.0 & 15.2 & 18.7 \\
  & Proposal Adjustment (PA)& 4.4 & 46.0 & 3.8 & 2.2 & 4.1 & 2.6 & 2.7 & 2.2 & 17.9 & 22.1 \\
  & Human-to-Object (HTO)& 3.8 & 45.7 & 4.4 & 2.4 & 4.5 & 2.7 & 3.0 & 2.4 & 19.1 & 23.4 \\
  & Density Estimation (DE) & \textbf{20.2} & 45.9 & 4.3 & 2.4 & 4.4 & 2.7 & \textbf{6.6} & \textbf{5.4} & 25.2 & 30.2 \\
  & Shortest Distance (SD) & 3.9 & 45.5 & 3.5 & 2.1 & 3.7 & 2.5 & 3.3 & 2.6 & 19.6 & 23.9 \\ 
  & HeatMap (HM) & 4.3 & 45.9 & 4.4 & 2.4 & 4.6 & 2.6 & 2.8 & 2.2 & 18.6 & 22.9 \\
  & Box Regression (BR) & 4.3 & 45.9 & 4.2 & 2.4 & 4.3 & 2.7 & 2.0 & 1.6 & 13.1 & 16.0 \\
  \cline{2-12}
  & HPN-P & 5.0 & 45.7 & 4.3 & 2.6 & 4.5 & 2.8 & 3.2 & 2.5 & 21.5 & 26.5 \\
  & HPN-H & 5.0 & 45.7 & 4.3 & 2.7 & 4.5 & 3.0 & 3.0 & 2.4 & 20.5 & 25.3 \\
  & HPN-C & 4.6 & 45.9 & 4.4 & 2.4 & 4.6 & 2.8 & 3.7 & 2.9 & 24.2 & 29.7 \\
  & \textbf{HPN} & 11.1 & \textbf{47.0} & \textbf{5.0} & \textbf{2.9} & \textbf{5.2} & \textbf{3.4} & 4.0 & 3.2 & \textbf{26.4} & \textbf{32.0} \\
  \hline
\end{tabular}}
\vspace{-0.3cm} 
\caption{Results on DIO. HPN-P/H/C indicates the single branch in HPN without dynamic fusion.} 
\label{tab:main-result} 
\vspace{-0.4cm}
\end{table*} 

\subsection{Implementation Details}
\label{sec:implementation}
\noindent{\bf Human Tracking.} 
FairMOT~\cite{fairmot} and DeepSORT~\cite{deepsort} are adopted and frozen.
For DeepSORT, we use the open-sourced version with YOLO~\cite{yolo} and pre-trained weights. 
For FairMOT, its pre-trained detector performs unsatisfied because of the domain gap between pedestrian data and AVA~\cite{AVA}. So we adopt YOLO~\cite{yolo} as its detector and only use its feature extractor pre-trained on multi-dataset~\cite{crowdhuman,caltechpedestrain,citypersons,cuhk-sysu,prw,ethz,milan2016mot16}.

\noindent{\bf Interaction Detection.} 
SlowFast~\cite{feichtenhofer2019slowfast} model pre-trained on Kinetics-700~\cite{kinetics} from \cite{jiajun} is finetuned on our train set for 24 epochs. We use an SGD optimizer, an initial learning rate of 4e-3, cosine learning rate decay, and a batch size of 16. 
A simple FC-Sigmoid classifier is used to classify interactions.
In training, we finetune the object discovery modules HPN-C, HPN-H, and HPN-P for 7, 22, and 11 epochs respectively.

\noindent{\bf Object Discovery.} 
To prepare the proposals, we finetune a COCO~\cite{coco} pre-trained RPN on the train set with GT boxes for 35K iterations. An SGD optimizer, a learning rate of 8e-4, and a batch size of 24 are adopted. 
We jointly train the object module and interaction classifier together for 1 epoch for HPN.
 For DE  with DeepSORT and FairMOT, we set the threshold $\alpha$ as 0.2 and 0.5.  For proposal-based methods with DeepSORT, $\alpha$ is set as 0.02. For the other methods, $\alpha$ is set as 0.03 and 0.4 for DeepSORT and FairMOT.
For baselines, we use the same learning rate and batch size as interaction detection. 
For single branch HPN-P/H/C, the heatmap filtering thresholds are all 0.6. 
For HPN, the thresholds for small, medium, and large objects are 0.7, 0.6, and 0.5 respectively. 
In the long-term module, $\epsilon$ is 0.1. 
The number of proposals $N_p$ in HPN-C is 75. 
These thresholds are obtained by grid search on val set.

\subsection{Results}
We report the results of HPN and baselines in Tab.~\ref{tab:main-result}.

\noindent{\bf Human Tracking.} 
For MOTA, DeepSORT largely surpasses FairMOT due to its person ID stability.
Since we only evaluate the individual seconds, the gap is more obvious. However, the IDF1 of FairMOT is slightly higher, which means the overall Re-ID accuracy of FairMOT is better.
Compared to FairMOT, the performances of baselines with DeepSORT is more robust, due to the strong tracking stability of DeepSORT.
Surprisingly, DE achieves impressive performance. Probably because it uses the proposal location probability to adjust the interaction score (Sec.~\ref{sec:baseline}), which can suppress the human-object pairs with low correlations and decrease the low-grade tracklets.
Relatively, HPN also shows decent improvements upon the other baselines and outperforms all methods on IDF1 of FairMOT.

\noindent{\bf Interaction Detection.}
Since we use the same Slowfast backbone and human tracking inputs, the performance variance of all methods is marginal.
Given FairMOT, HPN outperforms all baselines with the help of object and interaction joint training.
Meanwhile, with DeepSORT, DE achieves decent performance as it also considers two entangled tasks together.
But it performs worse than HPN with FairMOT, possibly because the person ID is unstable. 

\noindent{\bf Object Discovery.} 
HPN effectively captures the essential video cues and achieves great improvements under loose criterion, \eg, 2.5\% and 4.0\% improvements upon DE (DeepSORT).
But under strict criterion considering the quality of human tracking and interaction detection, DE is better.
Meanwhile, HTO also shows superiority over the other baselines.
These indicate that heatmap/proposal-based methods are promising for object discovery. But regression-based methods may lack the ability to locate these objects.
If only keeping one branch in HPN (HPN-P/H/C in Tab.~\ref{tab:main-result}, the context branch performs best. And they show obvious complementary properties, thus resulting in great improvement after the dynamic fusion.
Moreover, all methods perform unsatisfied under strict criterion because of the poor performances of human tracking and interaction detection. 
Hence, DIO poses a great challenge to vision systems and requires more studies on three sub-tasks.

\begin{figure}
    \centering
    \vspace{-0.3cm}
    \includegraphics[width=0.9\linewidth]{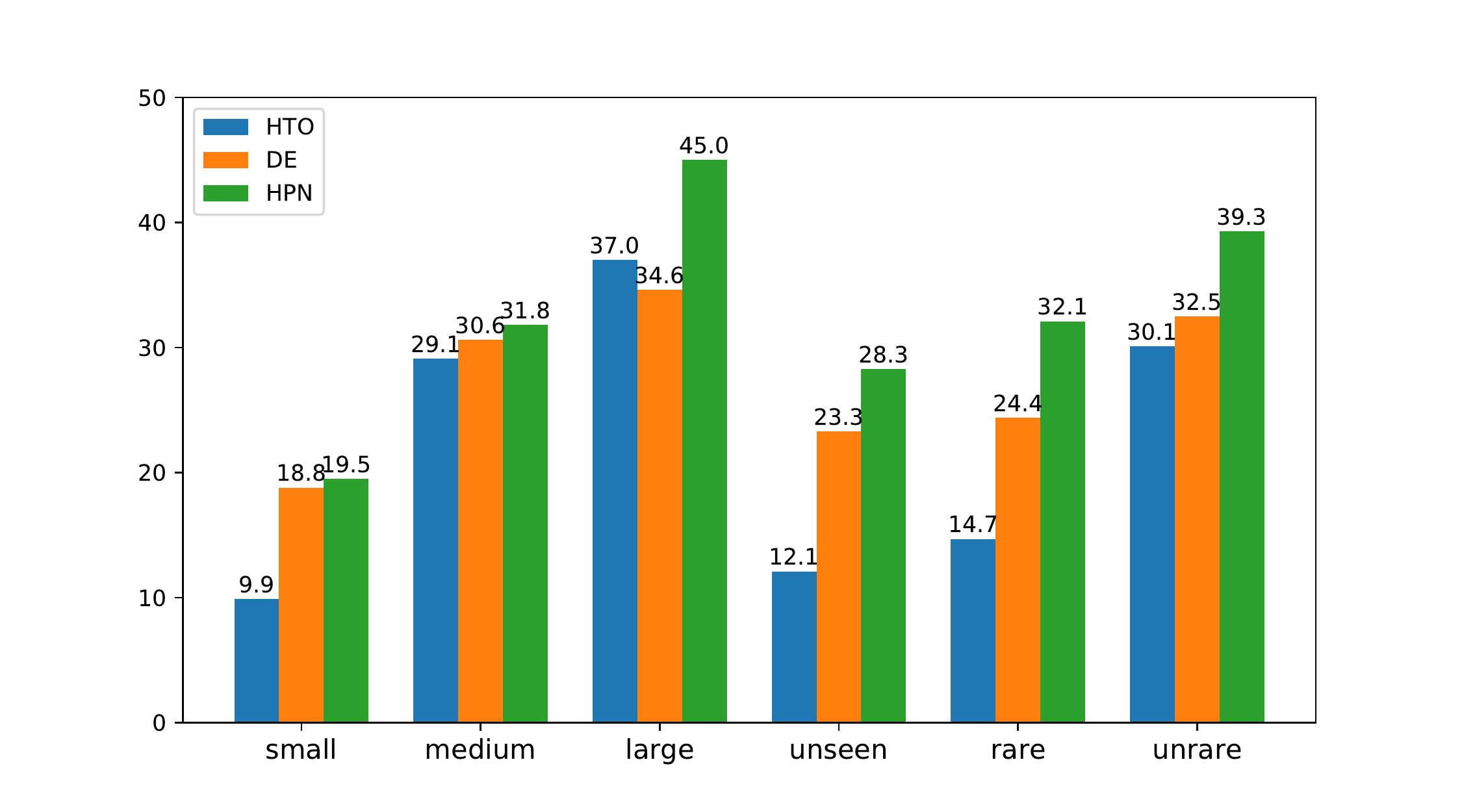}
    \vspace{-0.2cm}
    \caption{Object discovery analysis~(DeepSORT, loose 3D).}
    \label{fig:results} 
\vspace{-10px}
\end{figure}

\noindent{\bf Visualizations.} 
In Fig.~\ref{fig:results}, we analyze the object discovery results in different views, such as rare/unrare/unseen objects, size, \etc, where HPN shows superiority.

\subsection{Ablation Study}
We randomly choose 50 videos from the train set as val set for the ablation study on interacted object discovery.
HPN-C is chosen as the baseline, which achieves the best \textbf{32.2} mIoU (loose 3D) in the default setting, except for dynamic fusion. 
Detailed table please refer to our supplementary.

\textbf{(1) Losses}: replacing the BCE loss for regression with L1, L2 losses causes obvious degradations (\textbf{26.3}, \textbf{29.2} mIoU).

\textbf{(2) Fusion}: in three policies, dynamic fusion gets the best \textbf{35.9} mIoU, whilst equal fusion and prediction selection get \textbf{32.9} and \textbf{35.2} mIoU.

\textbf{(3) Condition}: in default, we translate the conditions into attentions, here we directly concatenate the conditioned feature with the perspective feature but perform worse (\textbf{31.1} mIoU).

\textbf{(4) Proposal}: we change the proposal numbers to 50 and 100 (\textbf{32.1, 32.1} mIoU) in the context STC map, but are all worse than 75.
\textbf{(5) Long-term Range}: we extend the long-term offset to 3 and 4 (\textbf{26.1}, \textbf{26.3} mIoU), but they are all worse than 2 (\textbf{27.9} mIoU).

\section{Conclusion}
In this work, we propose a novel ST-HOI learning task and construct a corresponding benchmark DIO. It contains 1,000+ object classes in 290K frame-level HOI triplets. 
To tackle this challenging task, we propose a hierarchical-probe network (HPN) and achieve decent results. 
However, DIO remains challenging, even after adopting state-of-the-art tools and methods. We believe it would inspire a new line of research on deeper activity understanding.

{\small
\bibliographystyle{ieee_fullname}
\bibliography{egbib}
}

\clearpage
\appendix

The contents of this supplementary material are:

Sec.~\ref{sec:characteristics}: Characteristics of DIO.

Sec.~\ref{sec:model-structure}: Detailed model structure.

Sec.~\ref{sec:baseline-details}: Detailed implementation of the baselines.

Sec.~\ref{sec:detailed-ablation}: Detailed ablation study.

Sec.~\ref{sec:visualization-results}: Visualized results of object discovery.

Sec.~\ref{sec:gt-input}: Testing results with GT as inputs.

\section{More characteristics of DIO}
\label{sec:characteristics}

\subsection{Video Selection for DIO Test Set}
\label{sec:video-select}
To make DIO challenging and practical, we construct its test set via seeing video selection as a \textbf{multi-objective integer programming} problem. 

\textbf{First}, given the video number $N$, interaction class number $N_a$, object class number $N_o$ and GT object location heatmap size $N_h$ (the original size of AVA~\cite{AVA} frames is resized to the size of the heatmap, and here we use a 1D vector to represent the values of original 2D heatmap) in AVA train-val sets, we define a binary variable $x_{i}\in\{0, 1\}, 1\le i \le N$ for each video to indicate whether to choose it or not for our test set. We restrict the sum of $x_{i}$ to the number of videos in the test set ($N_{t}$) according to a certain split ratio.

\textbf{Second}, for video $i$, we calculate its distributions of interaction class $\mathbf{a_i}\in \mathbb{N}^{N_a}$ (a set of interaction class frequencies), object class $\mathbf{o_i}\in \mathbb{N}^{N_o}$ (a set of object class frequencies), and object location GT heatmap $\mathbf{c_i}\in \mathbb{N}^{N_h}$. 
Each $x_{i}$ is multiplied to $\mathbf{a_i}$, $\mathbf{o_i}$ and $\mathbf{c_i}$ individually, then we add them up respectively to obtain three total distributions $\sum_{i=1}^N\mathbf{a_i}x_i$, $\sum_{i=1}^N\mathbf{o_i}x_i$, and $\sum_{i=1}^N\mathbf{c_i}x_i$ for all videos.

\textbf{Finally}, we want the test set to contain as many as possible interactions, object classes, and diverse object locations to fully evaluate the models. 
To this end, we calculate the variances $Var\left(\sum\limits_{i=1}^N\mathbf{a_i}x_i\right)$ and $Var\left(\sum\limits_{i=1}^N\mathbf{o_i}x_i\right)$ of interaction and object class distributions, use the variances as minimization objectives to search the suitable videos with the highest varieties of interaction and object classes. 
Moreover, we find that many objects are located at the half bottom of frames. Thus, to increase the variety of object location, we restrict the distribution of the top half part of heatmaps $\sum\limits_{i=1}^N\sum\limits_{k=1}^{N_h/2}\mathbf{c_{i, k}}x_i$ to a given threshold $\gamma$.
Additionally, to preserve the frequencies of some interaction classes from degrading to zero, we also add external restrictions on $\mathbf{a_i}$ with a threshold $\alpha_j$ for each interaction class $j$.  
The final programming problem to be solved is
\begin{align*}
&\min\quad z=Var\left(\sum\limits_{i=1}^N\mathbf{a_i}x_i\right)+Var\left(\sum\limits_{i=1}^N\mathbf{o_i}x_i\right) \\
& \begin{array}{c@{\quad}c}
s.t.&x_i\in\left\{0, 1\right\}, \quad1\le i \le N\\
    &\sum\limits_{i=1}^Nx_i=N_t,\\
    &\sum\limits_{i=1}^N\mathbf{a_{i, j}}x_i\geq\alpha_j,\\
    &\sum\limits_{i=1}^N\sum\limits_{k=1}^{N_h/2}\mathbf{c_{i, k}}x_i\geq\gamma.\\
\end{array}
\end{align*}
At last, the results are used to select the videos for our test split.

\subsection{Statistics of Data Split}
The detailed statistics of the data split are shown in Tab.~\ref{tab:dataset table}.

\begin{table}[]
\centering
\resizebox{0.6\linewidth}{!}{
\begin{tabular}{c|c|c|c}
\hline
Split & Box & Tracklet & Frame\\
\hline
train & 234K & 104K & 102K\\
\hline
test & 56K & 22K & 29K\\
\hline
all & 290K & 126K & 131K\\
\hline
\end{tabular}}
\caption{Statistics of data split.}
\label{tab:dataset table}
\end{table}

\begin{table*}[t]
\begin{center}
\resizebox{0.95\linewidth}{!}{
\begin{tabular}{|c|c|c|c|c|c|c|c|}
\hline
Action Id&Action Class&Action Id&Action Class&Action Id&Action Class&Action Id&Action Class \\ 
\hline
1 & jump/leap & 14 & dress/put on clothing & 27 & paint&40 & shoot\\
2 & lie/sleep & 15 & drink & 28 & play board game&41 & shovel\\
3 & sit & 16 & drive & 29 & play musical instrument&42 & smoke\\
4 & answer phone & 17 & eat & 30 & play with pets&43 & stir\\
5 & brush teeth & 18 & enter & 31 & point to&44 & take a photo\\
6 & carry/hold & 19 & exit & 32 & press&45 & text on/look at a cellphone\\
7 & catch & 20 & extract & 33 & pull&46 & throw\\
8 & chop & 21 & fishing & 34 & push&47 & touch\\
9 & clink glass & 22 & hit & 35 & put down&48 & turn\\
10 & close & 23 & kick & 36 & read&49 & watch\\
11 & cook & 24 & lift/pick up & 37 & ride&50 & work on a computer\\
12 & cut & 25 & listen & 38 & row boat&51 & write\\
13 & dig & 26 & open & 39 & sail boat& & \\
\hline
\end{tabular}}
\end{center}
\caption{Interaction class list of DIO.}
\label{tab:action list}
\end{table*}

\subsection{Data Samples}

\begin{figure*}
    \centering
    \includegraphics[width=0.95\linewidth]{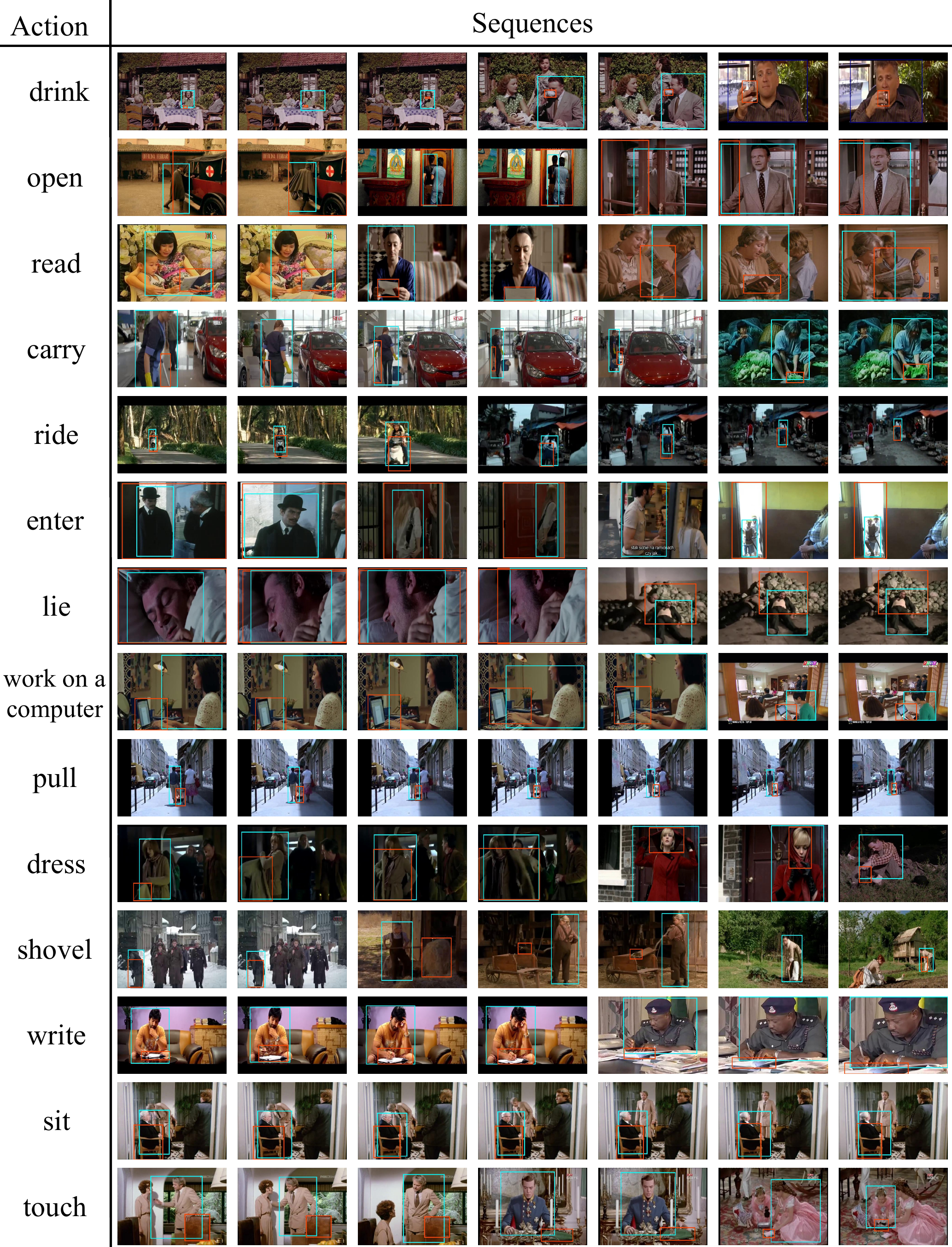}
    \caption{Data samples and their ST-HOI labels.}
    \label{fig:Data Sample}
\end{figure*}

Some data samples of DIO are shown in Fig.~\ref{fig:Data Sample}.
Human and object GT boxes are in blue and red respectively. 

\begin{algorithm}[t]
\SetAlgoLined
\KwIn{object class list $W$=$\{w_1, w_2, ...\}$}
\KwOut{cluster list $C$=$\{C_1, C_2, ...\}$}
 Initialize empty cluster list $C$\;
 \For{$i$=1:$|W|$}{
   \eIf{$|C|>0$}{
     \For{$j$=1:$|C|$}{
        Get $\hat{w}_j \in C_j$ with highest WordNet level\;
        \eIf {WordNet has path between ($w_i$, $\hat{w}_j$)}{
            Add $w_i$ to $C_j$\;
        }   
        {
            Add $w_i$ as a new cluster to $C$\;
        }
     }
   }{
    Add $w_i$ as a new cluster to $C$\;
  }
  $i$++\;
 }
 \caption{Clustering object classes.} 
 \label{alg:clustering}
\end{algorithm}

\subsection{Interaction List}
The detailed interaction classes are listed in Tab.~\ref{tab:action list}.

\subsection{Object Class Taxonomy}
To deal with the diversity of object class annotations in DIO, following EPIC-Kitchens~\cite{epic-kitchen}, we use WordNet~\cite{wordnet} to construct an object class tree. The detailed procedure is as follows:
\begin{itemize}
\item First, with the annotated object class list $W$=$\{w_1, w_2, ...\}$, we follow the clustering procedure of Algorithm \ref{alg:clustering} to build a cluster list $C$.
\item Then, we find some object classes are wrongly clustered due to the polysemy. 
For example, the first explanation of ``banana" in WordNet is a kind of ``herb", instead of ``fruit". For these classes, we manually remove them from $C$, correct their explanations and add them to $C$ as unique clusters.
\item Finally, we follow Algorithm \ref{alg:class_tree} to construct the object class tree with the clusters from $C$ and correct the ambiguous class names.
\end{itemize}

\begin{algorithm}[t]
\SetAlgoLined
\KwIn{cluster list $C$=$\{C_1, C_2, ...\}$}
\KwOut{object class tree $T$}
\SetKw{Break}{break}
\SetKwFunction{FMain}{ConstructTree}
\SetKwProg{Fn}{Function}{:}{}
\Fn{\FMain{$C_i$}}{
        \tcp{Construct object class tree $T$ from cluster $C_i$.}
        Initialize $T$ from the first word $w_1$ of $C_i$\;
        \For{$j$=2:$|C_i|$}{
            Get the $j$-th word $w_{i,j}$ of cluster $C_i$\;
            Get the node $T_k \in T$ with the shortest path between ($w_{i,j}$, $T_k$) in WordNet\;
            Add $w_{i,j}$ to $T_k$\;
         }
        \textbf{return} $T$
}
\SetKwFunction{FMain}{CombineTree}
\SetKwProg{Fn}{Function}{:}{}
\Fn{\FMain{$T_x$, $T_y$}}{
        \tcp{Combine object class tree $T_x$ and $T_y$.}
        Find root nodes $R_x$ and $R_y$ of $T_x$ and $T_y$\;
        Find closest common parent $R_{xy}$ of $R_x$ and $R_y$ in WordNet\;
        Add $R_x$ and $R_y$ to the children of $R_{xy}$\;
        Construct new class tree $T_{xy}$ from $R_{xy}$\;
        \textbf{return} $T_{xy}$
}
 Initialize object class tree $T$=ConstructTree($C_1$)\;
 \For{$i$=2:$|C|$}{
   $T_i$ = ConstructTree($C_i$)\;
   $T$ = CombineTree($T$, $T_i$)\;
 }
 \caption{Constructing object class tree.} 
 \label{alg:class_tree}
\end{algorithm}

\begin{figure*}[htbp]
    \centering
    \includegraphics[width=1\linewidth]{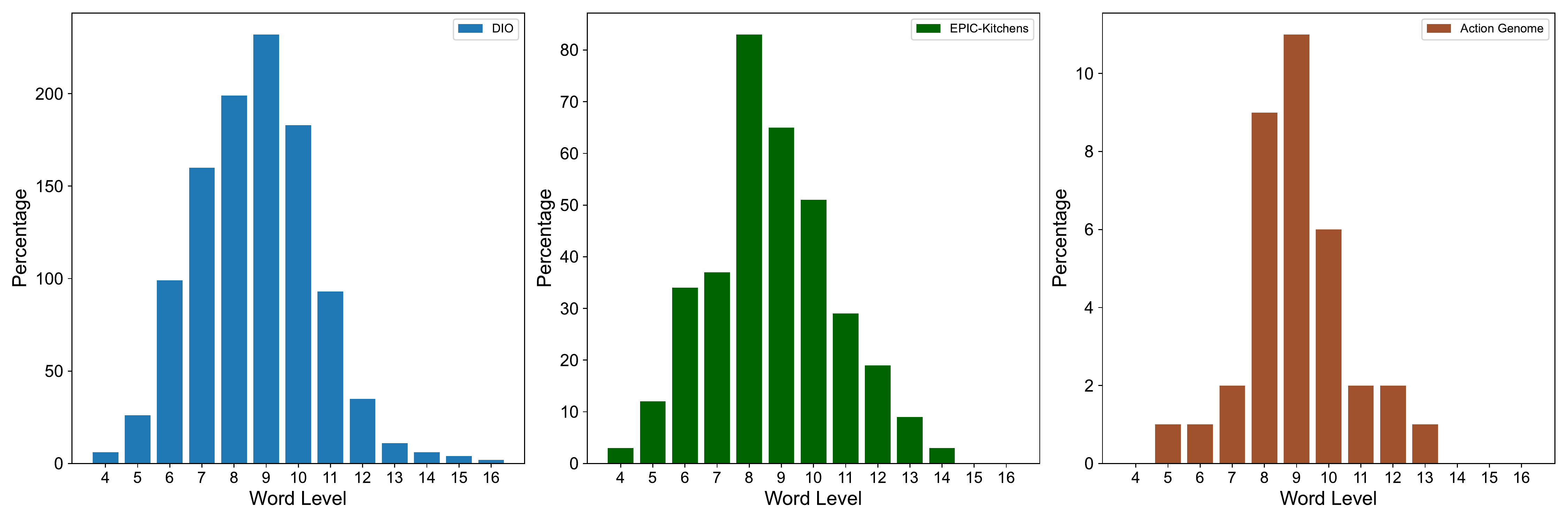}
    \caption{Comparison of object taxonomy between the datasets.}
    \label{fig:taxonomy}
\end{figure*}

Following the same procedure, we compare the class hierarchies, i.e., the ratio of different word levels in WordNet of DIO with the mainstream video datasets EPIC-Kitchens~\cite{epic-kitchen} and Action Genome~\cite{ji2020action}. The results in Fig.~\ref{fig:taxonomy} show that our DIO is more complex in classes and thus more challenging.

\subsection{Statistics of Action, Object, and Tracklet Length}
\begin{figure}[t]
    \centering
    \vspace{-0.5cm}
    \includegraphics[width=1\linewidth]{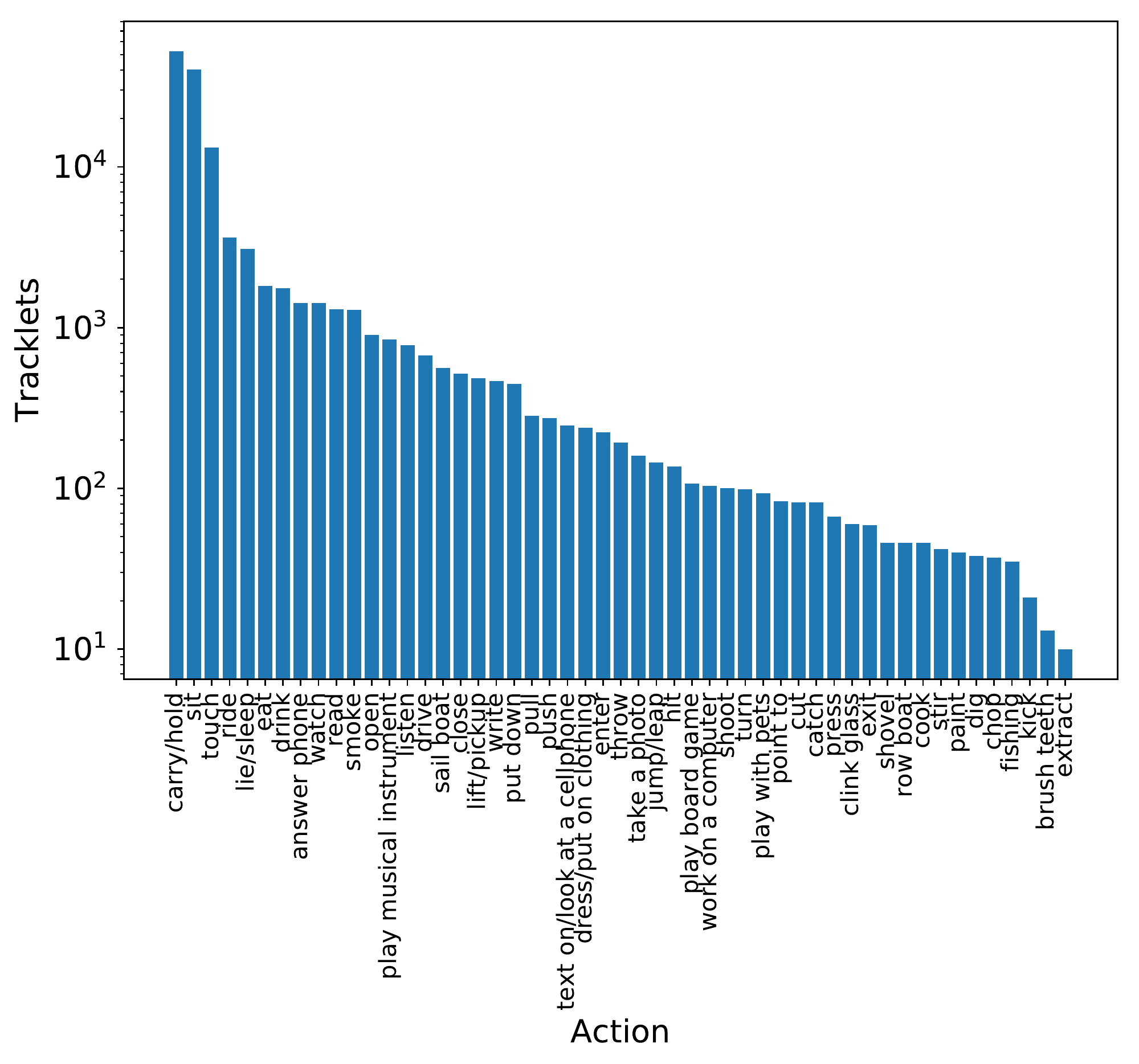}
    \vspace{-0.7cm}
    \caption{The distribution of tracklet number per action.}
    \label{fig:action}
\end{figure}

\begin{figure}[t]
    \centering
    \vspace{-0.5cm}
    \includegraphics[width=1\linewidth]{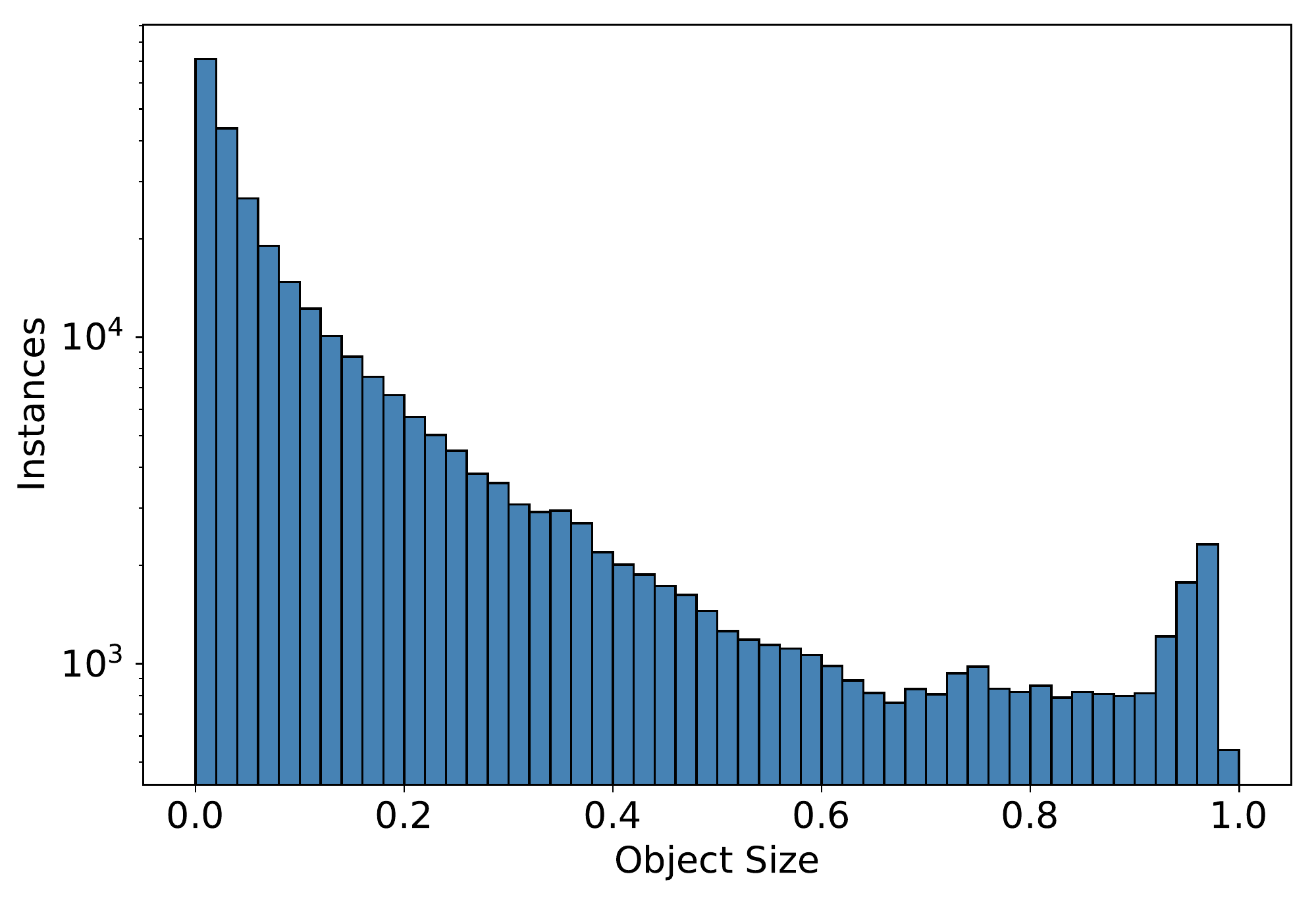}
    \vspace{-0.7cm}
    \caption{The distribution of normalized object size.}
    \label{fig:object}
\end{figure}

\begin{figure}[t]
    \centering
    \vspace{-0.5cm}
    \includegraphics[width=1\linewidth]{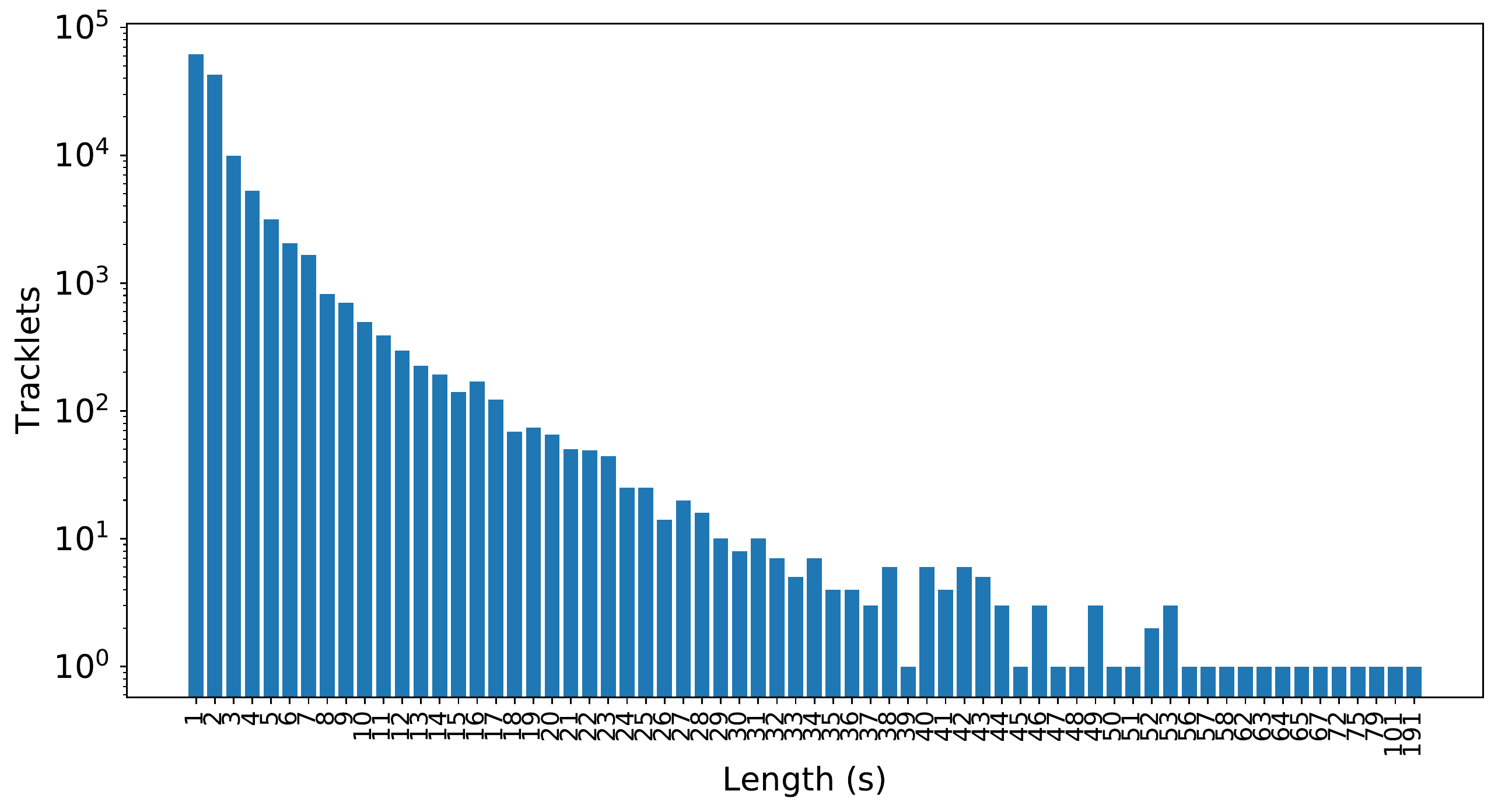}
    \vspace{-0.7cm}
    \caption{The distribution of tracklet length.}
    \label{fig:length}
\end{figure}

We also provide the distribution of action, object and tracklet length of DIO in Fig.~\ref{fig:action}-\ref{fig:length}.


\subsection{Video Clip Duration}
\begin{figure}[htbp]
    \centering
    \vspace{-9px} 
    \includegraphics[width=0.7\linewidth]{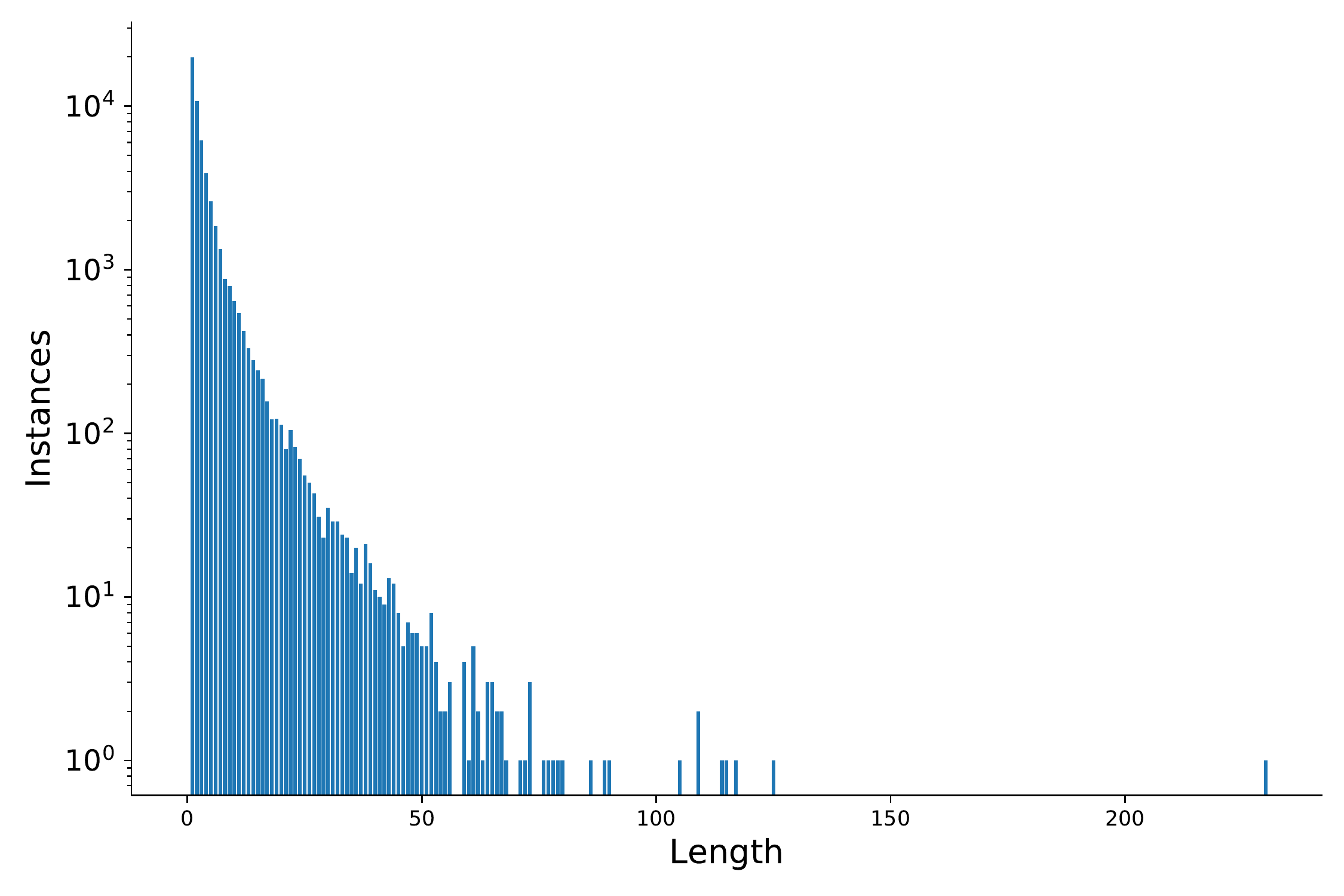}
    \caption{The statistics of temporal duration (second) in DIO.}
    \label{fig:duration}
\end{figure}

The statistics are shown in Fig.~\ref{fig:duration}.

\section{Detailed Model Structure}
\label{sec:model-structure}
\textbf{Encoder-Decoder.} 
For each branch of HPN, we adopt an encoder-decoder structure.
The detailed structure is presented in Tab.~\ref{tab:encoder_decoders}.

\begin{table}[htbp]
    \centering
    \resizebox{0.9\linewidth}{!}{
    \begin{tabular}{c|c|c}
    \hline
      Stage   & Layers & Output Size \\
    \hline
      \multicolumn{2}{c|}{Input Layer} & $C_{in} \times 56^2$ \\
      \hline
      \multirow{3}{*}{Encoder}& Conv, (64, 3, 2, 1) & $64 \times 28^2$ \\
       & Conv, (128, 3, 2, 1)& $128 \times 14^2$\\
       & Conv, (256, 3, 2, 1)& $256 \times 7^2$\\
      \hline
      \multirow{5}{*}{Decoder}& Deconv, (256, 3, 2, 1) & $256 \times 13^2$\\
      & Deconv, (128, 3, 2, 1) & $128 \times 25^2$\\
      & Deconv, (64, 3, 2, 1) & $64 \times 49^2$\\
      & Deconv, (32, 3, 2, 1) & $32 \times 97^2$\\
      & Deconv, (16, 3, 2, 1) & $16 \times 193^2$\\
      \hline
      \multirow{2}{*}{Regression Head} & Conv, (51, 1, 1, 0) & $51\times 193^2 $\\
                                    & Sigmoid & $51 \times 193^2$ \\
      \hline
    \end{tabular}}
    \caption{The detailed structure of the encoder-decoder used in our model. The layers are presented in (channels, kernel size, stride, and padding) format. $C_{in}=20, 2, 77$ for the part, human, and context branches of HPN individually.}
    \label{tab:encoder_decoders}
\end{table}

\textbf{Condition Module.} 
The condition module of HPN is detailed in Tab.~\ref{tab:condition}.
The inputs for the human part and human branches are both human appearance feature $f^{k}_h$ and verb vector $f^i_v$.
For the context branch, we additionally input the whole frame visual feature $f^k_c$ with $f^{k}_h, f^i_v$. The input size of our condition module is unified 4,608, and we use a zero vector to mask the context part of the visual condition for the human part and human branches.

\begin{table}[htbp]
    \centering
    \resizebox{0.95\linewidth}{!}{
    \begin{tabular}{c|c|c}
    \hline
      Stage   & Layers & Output Size \\
    \hline
      \multicolumn{2}{c|}{Visual Input Layer} & $4608 \times 7^2$ \\
      \hline
      \multirow{3}{*}{Visual Condition}& Conv, (1024, 1, 1, 0) & $1024 \times 7^2$ \\
       & Conv, (1024, 1, 1, 0) & $1024 \times 7^2$ \\
       & Global Average Pooling & $1024$ \\
      \hline\hline
      \multicolumn{2}{c|}{Semantic Input Layer} & $768$ \\
      \hline
      Semantic Condition & FC, 392 & $392$ \\
      \hline\hline
      \multicolumn{2}{c|}{V+S Concatenation} & $1416$ \\
      \hline
      \multirow{3}{*}{Condition Decoder}& FC, 256 & $256$\\
      & FC, 256 & $256$\\
      & Sigmoid & $256$ \\
      \hline
    \end{tabular}}
    \caption{The detailed structure of condition module. The convolutional layers are presented in (channels, kernel size, stride, and padding) format. For fully-connected layers, we only present their output dimensions. V+S means visual and semantic. }
    \label{tab:condition}
\end{table}

\textbf{Keypoint maps reweighting.} 
For the reweighting of $M^{k}_P$, we first perform Global Average Pooling (GAP) on $f^{k}_h$ and concatenate it with $f^i_v$. Then, we input the concatenated conditions to an MLP-Sigmoids layer to generate 18 attention values to reweight the 18 keypoint maps. The detailed structure is presented in Tab.~\ref{tab:reweighting}.

\begin{table}[htbp]
    \centering
    \resizebox{0.75\linewidth}{!}{
    \begin{tabular}{c|c|c}
    \hline
         Stage   & Layers & Output Size \\
         \hline
         \multicolumn{2}{c|}{Visual Input Layer} & $2304 \times 7^2$ \\
         \multicolumn{2}{c|}{Global Average Pooling} & $2304$ \\
         \hline\hline
         \multicolumn{2}{c|}{Semantic Input Layer} & $768$ \\
         \hline
         Semantic Condition & FC, 392 & $392$ \\
         \hline\hline
         \multicolumn{2}{c|}{V+S Concatenation} & $2696$ \\
         \hline
         \multirow{3}{*}{Reweighting} & FC, 1024 & $1024$ \\
         & FC, 18 & $18$ \\
         & Sigmoid & $18$ \\
         \hline
    \end{tabular}}
    \caption{The detailed structure of keypoint maps reweighting. V+S means visual and semantic.}
    \label{tab:reweighting}
\end{table}
\textbf{Dynamic Fusion.}
For dynamic fusion, we concatenate $\hat{M^{k}_P}, \hat{M^{k}_H}, \hat{M^{k}_C}$ and input them into an FC-Softmax to generate dynamic fusion factors $\beta_P, \beta_H, \beta_C$, then $h^{ki}=\beta_Ph^{ki}_P + \beta_Hh^{ki}_H + \beta_Ch^{ki}_C$. The detailed structure is presented in Tab. \ref{tab:fusion}.

\begin{table}[htbp]
    \centering
    \resizebox{0.75\linewidth}{!}{
    \begin{tabular}{c|c|c}
    \hline
         Stage   & Layers & Output Size \\
         \hline
         \multicolumn{2}{c|}{Input Layer} & $768$ \\
         \hline
         \multirow{3}{*}{Dynamic Fusion} & FC, 256 & $256$ \\
         & FC, 3 & $3$ \\
         & Softmax & $3$ \\
         \hline
    \end{tabular}}
    \caption{The detailed structure of dynamic fusion.}
    \label{tab:fusion}
\end{table}

\textbf{Long-term Module.}
The long-term module used in our model is also an encoder-decoder structure, with a 3D non-local layer and a temporal pooling layer after the encoder. The detailed structure is presented in Tab. \ref{tab:long_term}.

\begin{table}[htbp]
    \centering
    \resizebox{\linewidth}{!}{
    \begin{tabular}{c|c|c}
    \hline
      Stage   & Layers & Output Size \\
    \hline
      \multicolumn{2}{c|}{Input Layer} & $(2+N_p) \times (2\tau+1) \times 56^2$ \\
      \hline
      \multirow{3}{*}{Encoder}& 3D Conv, (64, 3, (1, 2), 1) & $64 \times (2\tau+1) \times 28^2$ \\
       & 3D Conv, (128, 3, (1, 2), 1)& $128 \times (2\tau+1) \times 14^2$\\
       & 3D Conv, (256, 3, (1, 2), 1)& $256 \times (2\tau+1) \times 7^2$\\
      \hline
      \multicolumn{2}{c|}{3D Non-local} & $256 \times (2\tau+1) \times 7^2$ \\
      \hline
      \multicolumn{2}{c|}{Temporal Average Pooling} & $256 \times 7^2$ \\
      \hline
      \multirow{5}{*}{Decoder}& 2D Deconv, (256, 3, 2, 1) & $256 \times 13^2$\\
      & 2D Deconv, (128, 3, 2, 1) & $128 \times 25^2$\\
      & 2D Deconv, (64, 3, 2, 1) & $64 \times 49^2$\\
      & 2D Deconv, (32, 3, 2, 1) & $32 \times 97^2$\\
      & 2D Deconv, (16, 3, 2, 1) & $16 \times 193^2$\\
      \hline
      \multirow{2}{*}{Regression Head} & 2D Conv, (51, 1, 1, 0) & $51\times 193^2 $\\
                                    & Sigmoid & $51 \times 193^2$ \\
      \hline
    \end{tabular}}
    \caption{The detailed structure of the long-term module. For 2D convolution, the layers are presented in (channels, kernel size, stride, and padding) format. For 3D convolution, the format is (channels, kernel size, (temporal stride, spatial stride), and padding).}
    \label{tab:long_term}
\end{table}

\textbf{Object Size Classifier.} For the structure of the object size classifier, we use a two-layer MLP classifier with Softmax. The input of the classifier is the concatenation of $f_{c}^k$ and $f_{h}^k$. For the object of the $k$-th second and $i$-th interaction, it would predict its size class. The detailed structure is presented in Tab. \ref{tab:size_classifier}.

\begin{table}[t]
    \centering
    \resizebox{0.75\linewidth}{!}{
    \begin{tabular}{c|c|c}
    \hline
         Stage   & Layers & Output Size \\
         \hline
         \multicolumn{2}{c|}{Input Layer} & $4608$ \\
         \hline
         \multirow{4}{*}{Object Size Classifier} & FC, 1024 & $1024$ \\
         & FC, 153 & $153$ \\
         & Reshape & $51 \times 3$ \\
         & Softmax & $51 \times 3$ \\
         \hline
    \end{tabular}}
    \caption{The detailed structure of object size classifier.}
    \label{tab:size_classifier}
\end{table}

Moreover, we use ReLU as the activation function and BatchNorm with a momentum of 0.9 after each convolutional and fully-connected layer in the modules presented above. 

\section{Baseline Details}
\label{sec:baseline-details}
In this section, we supplement some details of the baselines adopted in the experiment.

\noindent{\bf Proposal Selection~\cite{tan2019learning}.} 
With the RPN~\cite{fasterrcnn} finetuned on our train set, we can obtain the proposals of each frame in an ST-HOI tracklet. Naturally, we can directly select one as the predicted object.
Given the detected human and proposals, we extract their features $f^k_h, f^k_p$ and the context feature $f^k_c$. Next, $f^k_h, f^k_p$ are concatenated with $f^k_c$ and fed into an MLP interaction classifier trained with BCE loss. The more accurate a proposal, the better interaction classification it should provide. Thus, for $a_i$, the proposal with the highest score (performing best on the classification of $a_i$) is selected.

\noindent{\bf Proposal + Tracking-By-Detection (TBD).} 
Tracking-By-Detection~\cite{zhang2008global} provides a temporal association paradigm to adjust the proposal location on each second. Following MOT methods~\cite{milan2016mot16,ristani2016performance,fairmot,braso2020learning,deepsort}, we use Hungarian algorithm to match the proposals of the nearest two frames, and the Euclid distance matrix between proposal shape vectors $\{x, y, w, h\}$ is used as the cost matrix.
The mean interaction score (similar to the baseline proposal selection) of all proposals within a proposal tracklet is used as the tracklet score, the one with the highest mean score is chosen as the prediction.

\noindent{\bf Proposal + Single Object Tracking (SOT).} 
Another way to adjust the proposal locations temporally is SOT. We adopt SiamRPN~\cite{siamrpn} with AlexNet backbone as the tracker, input the proposals of the first frame and keep tracking until the final frame. In this way, for each human tracklet we will get the proposal object tubes with the same number of object proposals in one second. For all proposal tracklets, we use the interaction scores of their first frames to select the object predictions. After choosing the object tracklet with the highest prediction score, we can generate a continuous prediction of ST-HOIs. To stabilize the tracking results, in each second of one tracklet, we choose one proposal with the highest IoU to tracked box and directly use it as the tracking result. 

\noindent{\bf Proposal Adjustment~\cite{chen2017learning}.} 
For the proposal with the highest objectness score, we learn an offset between proposal $p$ and GT box $o$, \ie, $\delta = \{\frac{x_o-x_p}{w_p}, \frac{y_o-y_p}{h_p}, \log{\frac{w_o}{w_p}}, \log{\frac{h_o}{h_p}}\}$.
Then, we concatenate $f^k_h$, $f^k_p$, and $f^k_c$ and input them to an MLP regression head with two FCs. Following object detection, we use a smooth L1 loss to train it. 

\noindent{\bf Human-to-Object~\cite{gkioxari2018detecting}.} 
Following~\cite{gkioxari2018detecting}, we adjust the human box to the object box via an offset. 
This is in line with our human perspective.
For example, when the person is \textit{sitting} or \textit{riding} an object, the location of this interacted object is often under the person. 
With $f^k_h$ and $f^k_c$, we use a regression head same with proposal adjustment to learn $\delta$.
We use the action feature as input and add the same regression head which we designed for proposal adjustment after the action backbone. Then, we calculate the ground-truth offset between the human and object, i.e., $\delta_{ho} = \{\frac{x_o-x_h}{w_h}, \frac{y_o-y_h}{h_h}, \log{\frac{w_o}{w_h}}, \log{\frac{h_o}{h_h}}\}$.
Smooth L1 loss is also employed.
In testing, we adjust the tracked human box to recover the object location from the predicted offset $\hat{\delta}$.
Another similar way is to learn the offsets between human pose key points and object box~\cite{yang2019activity}. But it performs worse than the box adjustment on val set, thus we report the former implementation.

\noindent{\bf Density estimation~\cite{gkioxari2018detecting}}: 
We can also use the predicted object location as a selection condition for proposals.
The original implementation of ~\cite{gkioxari2018detecting} sees object discovery as density estimation. Given the offsets $\delta_{p|h}$ and $\delta_{o|h}$ from our method, density estimation use a Gaussian function to estimate the likelihood $p_o = e^{-{||\delta_{p|h}-\delta_{o|h}||}^2/2\sigma^2}$ that object locates in the predicted box. Compared to the inference of HPN, density estimation introduces a multiplication between $p_o$ and interaction score $S_a^{k}$, to take both action and object location into account.

\noindent{\bf Heatmap~\cite{kim2020tell}.} 
Recent anchor-free object detectors~\cite{centernet,cornernet,fcos} have shown the capability of objectiveness heatmap to predict the object location distribution. 
We can also use heatmap estimation to localize objects similar to \cite{kim2020tell} and HPN.
Differently, we direct input the video cues to the model to estimate the heatmap~\cite{kim2020tell}.
With the human skeleton map, we first use two convolution layers to extract its feature. Then we concatenate it with $f^k_v$ and $f^k_h$ and input them into a decoder to estimate the heatmap. In testing, the normalization, and filtering with the same threshold of HPN are operated for the heatmap. And the tightest box of the binary mask is adopted as the predicted box.

\noindent{\bf Box Regression.} 
A pure regression without the help of proposals is also considered.
We input the concatenated $f^k_h$ and $f^k_c$ into two FCs with a subsequent ReLU activation and output $\{x, y, w, h\}$ as the location. This simple model is trained with a smooth L1 loss between $\{x, y, w, h\}$ and the GT box.
For the test stage, we directly output the predicted object boxes through the network.

\begin{figure*}
    \centering
    \includegraphics[width=0.88\linewidth]{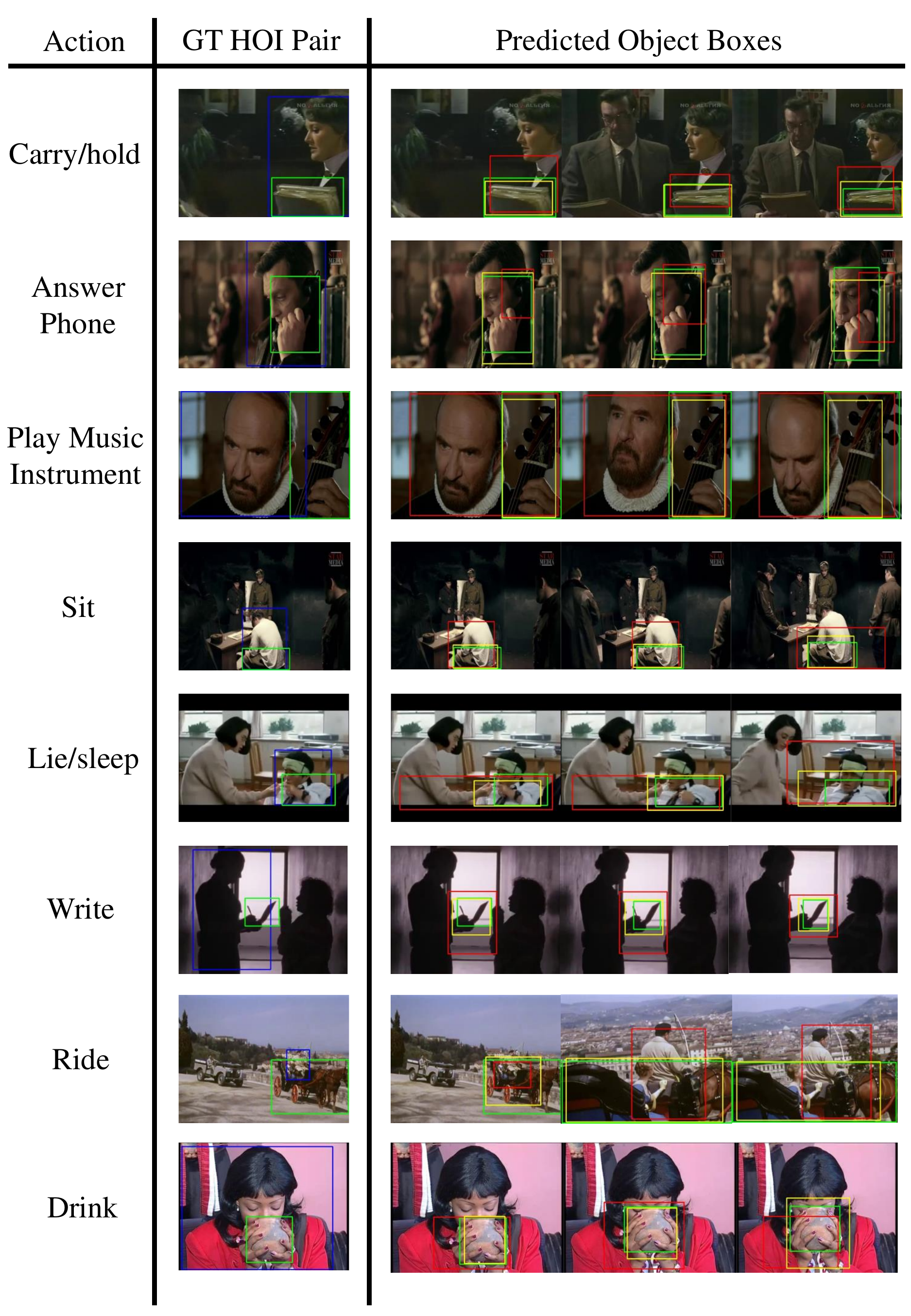}
    \vspace{-0.3cm}
    \caption{Visualizations of interacted object discovery. Boxes: human (\textcolor{blue}{blue}), GT (\textcolor[RGB]{0, 230, 0}{green}), HPN (\textcolor[RGB]{210, 210, 0}{yellow}), DE(\textcolor{red}{red}).} 
    \label{fig:results_supp} 
\vspace{-0.6cm} 
\end{figure*}

\section{Detailed Ablation Study}
\label{sec:detailed-ablation}
We compare the object discovery performance of HPN on the validation set with different losses, fusion policies, conditions, proposal numbers, and long-term ranges. 
We report the results in loose 2D and loose 3D metrics. 
The detailed results are shown in Tab.~\ref{tab:ablation_loss}-\ref{tab:ablation_long}.

\begin{table}[htbp]
    \centering
    \resizebox{0.65\linewidth}{!}{
    \begin{tabular}{c|c|c}
    \hline
         \multirow{2}{*}{Loss Function} & \multicolumn{2}{c}{mIoU(\%)} \\
         \cline{2-3}
          & Loose 2D & Loose 3D \\
         \hline
         \textbf{BCE} & \textbf{30.6} & \textbf{32.2} \\ 
         Smooth L1 & 25.0 & 26.3 \\ 
         L2 & 27.8 & 29.2 \\ 
         \hline
    \end{tabular}}
    \caption{The influence of different loss functions.}
    \label{tab:ablation_loss}
\end{table}

\begin{table}[htbp]
    \centering
    \resizebox{0.7\linewidth}{!}{
    \begin{tabular}{c|c|c}
    \hline
         \multirow{2}{*}{Fusing Policies} & \multicolumn{2}{c}{mIoU(\%)} \\
         \cline{2-3}
          & Loose 2D & Loose 3D \\
         \hline
         \textbf{Dynamic Fusing} & \textbf{34.2} & \textbf{35.9} \\
         Equal Late Fusion & 31.1 & 32.9 \\ 
         Prediction Selection & 33.6 & 35.2 \\ 
         \hline
    \end{tabular}}
    \caption{The influence of different fusing policies.}
    \label{tab:ablation_fusing}
\end{table}

\begin{table}[htbp]
    \centering
    \resizebox{0.7\linewidth}{!}{
    \begin{tabular}{c|c|c}
    \hline
         \multirow{2}{*}{Condition} & \multicolumn{2}{c}{mIoU(\%)} \\
         \cline{2-3}
          & Loose 2D & Loose 3D \\
         \hline
         \textbf{Attention} & \textbf{30.6} & \textbf{32.2} \\
         Concatenation & 29.6 & 31.1 \\
         \hline
    \end{tabular}}
    \caption{The influence of different condition translations.}
    \label{tab:ablation_condition}
\end{table}

\begin{table}[htbp]
    \centering
    \resizebox{0.75\linewidth}{!}{
    \begin{tabular}{c|c|c}
    \hline
         \multirow{2}{*}{Proposal Number $N_p$} & \multicolumn{2}{c}{mIoU(\%)} \\
         \cline{2-3}
          & Loose 2D & Loose 3D \\
         \hline
         $N_p=50$ & 30.4 & 32.1\\ 
         $\mathbf{N_p=75}$ & \textbf{30.6} & \textbf{32.2} \\ 
         $N_p=100$ & 30.4 & 32.1 \\ 
         \hline
    \end{tabular}}
    \caption{The influence of different proposal numbers.}
    \label{tab:ablation_proposal}
\end{table}

\begin{table}[htbp]
    \centering
    \resizebox{0.75\linewidth}{!}{
    \begin{tabular}{c|c|c}
    \hline
         \multirow{2}{*}{Long-term Range $\tau$} & \multicolumn{2}{c}{mIoU(\%)} \\
         \cline{2-3}
          & Loose 2D & Loose 3D \\
         \hline
         $\mathbf{\tau=2}$ & \textbf{26.7} &  \textbf{27.9}\\ 
         $\tau=3$ & 25.0 & 26.1\\ 
         $\tau=4$ & 25.2 & 26.3\\ 
         \hline
    \end{tabular}}
    \caption{The influence of different long-term ranges. Note that we \textbf{only} evaluate the performance of the long-term module instead of the whole model in this experiment.}
    \label{tab:ablation_long}
\end{table}

\begin{table*}[t]
\begin{center}

\resizebox{0.75\linewidth}{!}{
\begin{tabular}{@{\extracolsep{4pt}}cc cc cc cc cc@{}}
\hline
  \multirow{3}{*}{Methods} & \multicolumn{2}{c}{Interaction Detection (mAP(\%))} & \multicolumn{4}{c}{Object Discovery (mIoU(\%))} \\ 
 \cline{2-3}\cline{4-7}
  & \multirow{2}{*}{2D} & \multirow{2}{*}{3D} & \multicolumn{2}{c}{Strict} & \multicolumn{2}{c}{Loose}\\ 
  
  \cline{4-5}\cline{6-7}\cline{8-9}
  & & & 2D & 3D & 2D & 3D \\
  \hline
  Proposal Selection (PS) & 18.1 & 13.9 & 8.4 & 8.0 & 20.9 & 22.2 \\
  Proposal + TBD (PTBD) & 18.1 & 13.8 & 8.8 & 8.4 & 21.8 & 23.3 \\
  Proposal + SOT (PSOT) & 18.2 & 13.9 & 8.2 & 7.8 & 20.9 & 22.3 \\
  Proposal Adjustment (PA) & 18.0 & 13.9 & 10.7 & 10.4 & 25.1 & 26.9 \\
  Human-to-Object (HTO) & 19.0 & 14.8 & 11.9 & 11.5 & 28.6 & 30.7 \\ 
  Density Estimation (DE) & 19.3 & 14.3 & 11.8 & 11.4 & 31.9 & 34.5 \\
  Shortest Distance (SD) & 17.9 & 13.7 & 11.1 & 10.7 & 27.5 & 29.5 \\
  HeatMap (HM) & 18.1 & 14.1 & 11.3 & 10.9 & 27.5 & 29.5 \\
  Box Regression (BR) & 17.5 & 13.7 & 8.4 & 8.1 & 19.2 & 20.6 \\ 
  \hline
  HPN-P & 18.6 & 14.7 & 10.9 & 10.5 & 30.8 & 33.1 \\
  HPN-H & 18.6 & 14.4 & 10.3 & 9.9 & 29.7 & 31.9 \\
  HPN-C & 19.0 & \textbf{14.9} & 12.8 & 12.4 & 33.8 & 36.2 \\
  \textbf{HPN} & \textbf{19.7} & 14.7 & \textbf{13.0} & \textbf{12.5} & \textbf{37.0} & \textbf{39.7} \\
  \hline
\end{tabular}}
\end{center}
\caption{ST-HOI learning results with \textbf{ground-truth human tracklets} as inputs.}
\label{tab:gt_box}
\end{table*}

\begin{table*}[t]
\centering
\resizebox{0.45\linewidth}{!}{
\begin{tabular}{@{\extracolsep{4pt}}cc cc cc cc cc@{}}
\hline
  Methods & Object Discovery (mIoU(\%)) \\ 
  \hline
  Proposal Selection (PS) & 23.0 \\
  Proposal + TBD (PTBD) & 23.9 \\
  Proposal + SOT (PSOT) & 23.2 \\
  Proposal Adjustment (PA) & 25.6 \\
  Human-to-Object (HTO) & 28.7 \\ 
  Density Estimation (DE) & 32.8 \\
  Shortest Distance (SD) & 27.8\\
  HeatMap (HM) & 28.2 \\
  Box Regression (BR) & 19.4\\ 
  \hline
  HPN-P & 31.8 \\
  HPN-H & 30.4 \\
  HPN-C & 34.6 \\
  \textbf{HPN} & \textbf{37.8} \\
  \hline
\end{tabular}}
\caption{ST-HOI learning results with \textbf{ground-truth human tracklets and action labels} as inputs.}
\label{tab:gt_box_action}
\end{table*}

\section{Visualized Object Discovery Results}
\label{sec:visualization-results}
In Fig.~\ref{fig:results_supp}, we compare predicted boxes with GT boxes on the test set given GT human tracklets. HPN robustly estimates reasonable object locations for various interactions. A video visualization is also provided in the \textbf{demo.mp4} file in the supplementary folder.

\section{Testing with Ground-truth Inputs}
\label{sec:gt-input}
We measure the \textbf{upper bound} of object discovery by inputting \textbf{ground-truth human boxes and action labels}. 
The results under two modes are reported: 
GT human boxes only (H only) and GT human boxes with action labels (H+A). 
The results are detailed in Tab.~\ref{tab:gt_box} and Tab.~\ref{tab:gt_box_action}. 
We can find that under both scenarios, HPN also shows great superiority compared to all baselines. 
Moreover, given the same GT human tracklets, the advantage of DE has been weakened.

\end{document}